\documentclass[preprint,12pt,authoryear]{elsarticle}
\usepackage[top=3cm,bottom=3cm, left=2.2cm, right=2.2cm]{geometry}
\usepackage{framed,multirow}
\usepackage{graphicx}
\usepackage{amssymb}
\usepackage{amsthm}
\usepackage{latexsym}

\usepackage{url}
\usepackage{xcolor}

\usepackage{hyperref}

\usepackage{amsmath}

\usepackage{comment}
\usepackage{float}
\usepackage{soul}

\definecolor{newcolor}{rgb}{.8,.349,.1}
\newtheorem{theorem}{Theorem}
\newtheorem{definition}{Definition}

\begin{document}

\begin{frontmatter}



\title{Scale-invariant brain morphometry:\\ application to sulcal depth}


\author{Maxime Dieudonn\'e$^a$, Guillaume Auzias$^{a\dagger}$, Julien Lef\`evre$^{a\dagger}$\\
\small$^\dagger$ GA and JL contributed equally to this work.\\[-1mm]
}
\affiliation{organization={Institut de Neurosciences de la Timone, UMR 7289, CNRS, Aix-Marseille Université},
            addressline={}, 
            city={Marseille},
            postcode={13005}, 
            state={},
            country={France}}

\begin{abstract}
The geometry of the human cortex is complex and highly variable, with interactions between brain size, cortical folding, and age well-documented in the literature. However, few studies have explored how global brain size influences morphometry features of the cortical surface derived from anatomical MRI. In this work, we focus on sulcal depth, an imaging phenotype that has gained attention in both basic research and clinical applications. We make key contributions to the field by: 1) providing the first quantitative analysis of the influence of brain size on sulcal depth measurements; 2) introducing a novel, scale-invariant method for sulcal depth estimation based on an original formalization of the problem; 3) presenting a validation framework and sharing our code and benchmark data with the community; and 4) demonstrating the biological relevance of our new sulcal depth measure using a large sample of 1,987 subjects spanning the developmental period from 26 weeks post-conception to adulthood.

\end{abstract}

\begin{graphicalabstract}
\includegraphics[width=\linewidth]{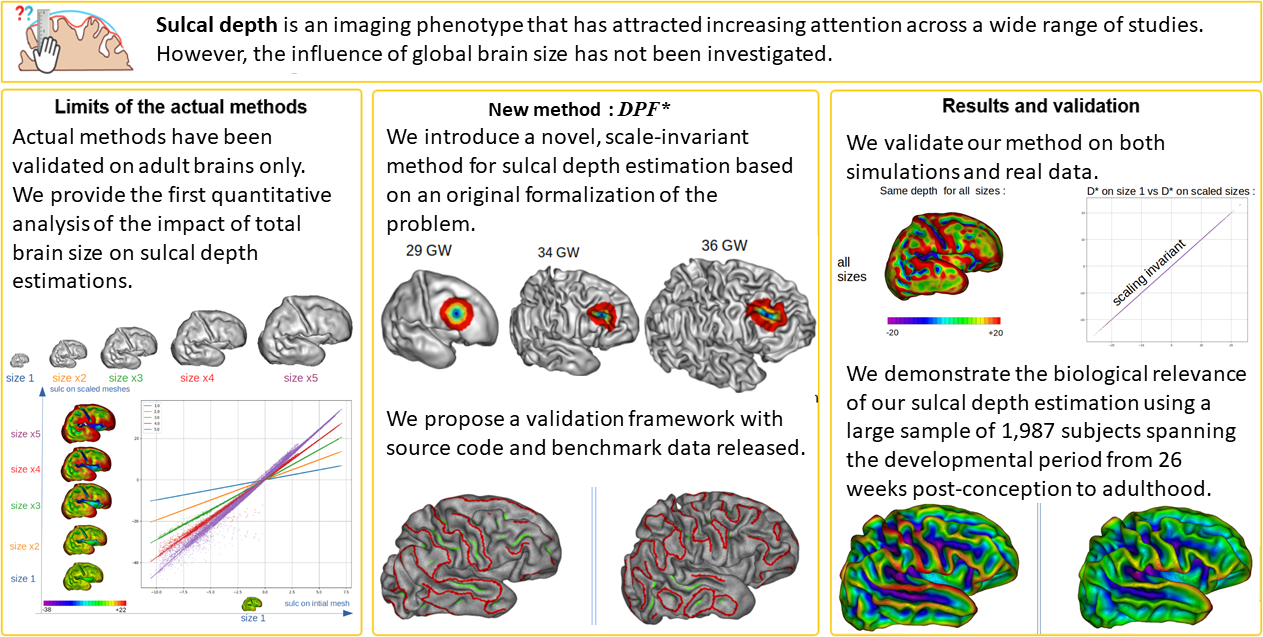}
\end{graphicalabstract}

\begin{highlights}
\item The first quantitative analysis of how brain size affects sulcal depth measurements.
\item A novel method method based on an original formalization of the problem.
\item A validation framework with code and benchmark data released.
\item Experiments on a large sample of 1,987 subjects spanning the developmental period.
\end{highlights}

\begin{keyword}
Magnetic resonance Imaging, \sep cortical surface, \sep sulcal depth


\end{keyword}

\end{frontmatter}

\newcommand{\comjul}[1]{\textcolor{magenta}{#1}}
\newcommand{\commax}[1]{\textcolor{BurntOrange}{#1}}
\newcommand{\comgui}[1]{\textcolor{teal}{#1}}

\section{Introduction}
\label{intro}
Sulcal depth is an imaging phenotype that has attracted increasing attention across a wide range of studies, from fundamental research to clinical applications. Key applications include inter-individual cortical surface registration and atlas construction \cite{dale_cortical_1999, lyttelton_unbiased_2007, robinson_multimodal_2018}, detection of abnormal brain morphology related to clinical conditions \cite{auzias_atypical_2014, dierker_analysis_2013}, and exploration of the anatomo-functional relationships \cite{weiner_mid-fusiform_2013}.
Furthermore, sulcal depth plays a crucial role in studying individual variations in cortical folding at various spatial scales, from sulci \cite{duchesnay_classification_2007} to sulcal basins \cite{auzias_deep_2015, im_brain_2008}, and even more specific geometrical structures such as plis-de-passages \cite{bodin_plis_2021, mangin_plis_2019} or gyral peaks \cite{zhang_gyral_2023}.
Despite its recognized importance for characterizing brain morphology, the sulcal depth is still underutilized compared to other widely used morphological descriptors, such as the gyrification index \cite{schaer_surface-based_2008, rabiei_local_2017}, curvature-related characteristics \cite{pienaar_methodology_2008}, and sulcus-based morphometry \cite{mangin_-vivo_2010}. This limited use can be largely attributed to the inconsistent definitions of sulcal depth used in different studies, which complicates a unified understanding of the literature. Moreover, current methods for estimating sulcal depth often overlook the potential influence of variations in global brain size.



\subsection{On the magnitude of variations in global brain size}
The inter-individual variations in global brain size has been extensively described in the literature.
The recent large-scale study \cite{bethlehem_brain_2022} aggregated morphometry measures extracted from a very large population spanning the entire life, using image processing pipelines such as Freesurfer \cite{dale_cortical_1999},  or the dHCP pipeline \cite{makropoulos_developing_2018}.
The authors applied advanced non-linear regression techniques to estimate the influence of the age on these morphometry measures.
From these curves representative of a large population, we can observe a ~4-fold increase in total brain volume between birth and adulthood.
In a recent work investigating brain maturation before birth using MRI acquired \textit{in utero}, \cite{studholme_motion_2020} reported a ~12-fold increase between the $18^{th}$ week of gestation and birth.
So regarding the human brain, variations related to brain maturation can reach an order of magnitude of ~48-fold.
Regarding inter-individual variations, we can also observe from \cite{bethlehem_brain_2022} a ~3-fold variation in total brain volume ($[0.5,1.5]$ L) across healthy adults brains.
A large body of publication investigated the putative sources of the observed variations across individuals, such as the effect of the sex \cite{diaz-caneja_sex_2021, fish_influences_2017}, the relationship between depth and cortical thickness and curvature \cite{demirci_systematic_2023}, or the developmental trajectory \cite{duan_individual_2020}.
On the side of cross-species comparisons, studies such as \cite{Van_essen_cerebral_2019} reported ~17-fold variation in total brain volume between mature macaques and mature humans.
See also \cite{demirci_cortical_2022} for an analysis of the variations across 12 primate species covering a wide range of sizes and forms.
Taking into account the variations in global brain size when comparing measures across individuals or populations is critical in most applications, either basic or clinically oriented.

\subsection{Features estimated locally are affected by global brain size}
It is also crucial to keep in mind that the relationship between global brain size and local measures of brain shape is not restricted to a linear interaction.
Studies such as \cite{reardon_normative_2018} characterized these complex interactions and reported regionally specific patterns of non linear areal scaling in the human cortex across healthy adults. 
Such interactions between size and shape (often termed allometry) are very common in biological systems \cite{goriely_mathematics_2017}.
More specifically, systematic interactions between brain size, cortical folding and age have been extensively reported in the literature, both before and after birth \cite{rodriguez-carranza_framework_2008, yun_temporal_2020, garcia_dynamic_2018, xu_spatiotemporal_2022, bozek_construction_2018}.
Such interactions require particular attention in the context of developmental diseases that affect both brain size and cortical folding \cite{kapellou_abnormal_2006, germanaud_simplified_2014}

\subsection{State-of-the-art of sulcal depth estimation techniques}
\label{sec:SOTA}

The most intuitive strategy consists in defining the sulcal depth at each point of a cortical surface by computing the distance between the spatial location of the point of interest and the closest point on the 'convex hull' of the brain.
Already with this intuitive definition, two main ambiguities emerge when moving from the concept to its implementation.
First, computing the 'convex hull' that implicitly corresponds to the 0-level (or reference-level) for the depth measure is not trivial. Indeed, a mathematically correct definition of the convex hull of the cortical surface would not follow closely most concave regions such as the orbital frontal region and the internal temporal region, resulting in a local over estimation of the sulcal depth compared to e.g. the external, more convex regions. The available implementations thus rely instead on a proxy computed by applying a morphological closing of the volumetric segmentation mask of the cortex, which is efficient but question the potential influence between the size of the closing operator and the global size of the brain of interest. 
Second, there is no unique way to compute the distance to the convex hull and several approaches have been proposed, such as the Euclidean distance in 3D \cite{van_essen_surface-based_2004}, the geodesic distance along the cortical surface \cite{cachia_generic_2003}, or an adaptive distance in 3D that follows the geometry of the surface \cite{yun_automated_2013}. We refer the readers interested in a detailed comparison across these types of approaches to \cite{yun_automated_2013} which provides both qualitative and quantitative comparisons.

A second strategy for estimating sulcal depth without considering the convex hull has been introduced in \cite{dale_cortical_1999}. We refer to this method as SULC in the following.
SULC is computed iteratively through an inflation process deforming an initial surface into a smooth, inflated one. 
Formally, the SULC estimation is obtained as:
\begin{equation} 
\label{eq:sulc}
SULC(i) = \int_{0}^{\infty}S_i(t) \cdot N_i(t)dt
\end{equation}
where $N_i(t)$ is the normal at $i$, $S_i(t)$ represents the infinitesimal deformation at vertex $i$ and time $t$, obtained through the gradient of an energy preserving local distances $d_{iv}=||e_{iv} ||$ where $e_{iv}$ is the vector between vertex $i$ and one of its neighbours $v$
\begin{equation} 
\label{eq:defor}
S_i(t) = \frac{1}{\#M} \sum_{v \sim i } e_{iv}(t) \Bigg(1 + \lambda \frac{d_{iv}(t)-d_{iv}(0)}{d_{iv}(t)} \Bigg)   
\end{equation}

In this approach, the reference (zero) level is defined implicitly.
Since the balance between the amount of gyri and sulci depends on the magnitude of gyrification, which itself is correlated with total brain size and/or age, variations in global brain size are expected to influence the local sulcal depth estimation in SULC.

A third strategy for estimating sulcal depth is based on two hypotheses relating sulcal depth and mean curvature: (1) the location of the extrema of mean curvature are a first approximation of the location of the extrema of sulcal depth and (2) the frequency ad amplitude of curvature and sulcal depth are closely related (\cite{luders_curvature-based_2006}, \cite{chung_deformation-based_2003}).

Finally, a specific method denoted as Depth Potential Function (DPF) has been introduced in \cite{boucher_depth_2009} as a combination of the second and third strategies. 
The DPF has been defined as the solution $D$ of the following regularised screened Poisson equation:
\begin{equation} 
\label{eq:DPF}
(-\Delta_M  + \alpha I)D = 2K
\end{equation}
where $\Delta_M$ is the Laplace-Beltrami operator of the surface $M$ (negative operator), $I$ is the identity operator, $K$ the mean curvature of $M$ and $\alpha$ a parameter.
Here also, the zero level is defined implicitly, and the potential influence of global brain size is not considered in this formal definition.

In addition to the specific limitations mentioned above, all the methods from the literature have been designed on adult brains only.
To the best of our knowledge, the potential influence of global brain size on the sulcal depth estimation obtained at each vertex of the cortical surface was not considered in the publication introducing these methods.
The interactions between global brain size and cortical folding magnitude has been explored rarely, with a few notable exceptions:
The excellent work \cite{pienaar_methodology_2008}, introduces the problem of normalization with respect to global brain size but focuses on curvature and not sulcal depth.
In \cite{meng_spatial_2014}, the authors investigate the evolution of locally deepest points in sulci (sulcal pits) during the first 2 years of life. In order to compensate for the variations in brain size across different subjects at different ages that affect the sulcal depth estimates, the authors propose to adapt the parameters of the sulcal pits extraction algorithm as follows: they perform linear regressions in order to approximate relationship between the parameters of the algorithm and respectively the total area maximum and maximal sulcal depth of the cortical surface. The evaluation of the effectiveness of this approach relies on the observed increase in the number of sulcal pits with age that is consistent across major sulci.This study illustrates the need for better formalization of the impact of global brain size on the depth estimation itself.
Indeed, in application studies, global brain size is usually considered as a confounding factor to be controlled as much as possible.
For instance, in \cite{natu_sulcal_2021} the authors investigate the relationship between functional activations and sulcal depth (using SULC) in the medial ventral temporal sulcus, in human children and adults, and macaques. The potential influence of global brain size is ruled-out by normalizing the sulcal depth estimates according to the deepest point within the sulcus of interest.
There is thus a clear need to explicitly investigate more specifically the problem of the intricate variations in global brain size and folding. 



\subsection{Contributions}
In the present work, we contribute to this field of research by 1) providing the first quantitative characterization of the influence of brain size on sulcal depth estimations; 2) introducing a new scale-invariant sulcal depth estimation method based on an original formalization of the problem; 3) proposing a validation framework and sharing with the community the benchmark data; 4) demonstrating the biological significance of our new sulcal depth measure using a large sample of 1,987 subjects, spanning the developmental period from 26 weeks post-conception to adulthood. 
The source code for this work is available at \href{https://github.com/maximedieudonne/DPF-star}{https://github.com/maximedieudonne/DPF-star}, and the benchmark data is available at \href{https://doi.org/10.5281/zenodo.14265228}{https://doi.org/10.5281/zenodo.14265228}.
This article extends our preliminary work \cite{dieudonne_scale-controlled_2024}.



\section{Method}

In this section, we propose a formalisation of the influence of the global size of a surface on any kind of functions defined on it (Sec.\ref{sec:scale_control}.
We pursue by exposing formally the constraints allowing us to define functions on surfaces for which we control the influence of global size. 
In Sec.\ref{sec:DPF_star}, we then focus on the specific case of functions dedicated to the estimation of sulcal depth from a cortical surface, and apply our definitions in order to introduce the normalised version of the $DPF$, that we call $DPF^*$.
Finally, in Sec.\ref{sec:alpha_setting}, we precise the influence of the only free parameter of $DPF^*$.

\subsection{Scale controlled family of functions on surfaces}
\label{sec:scale_control}

Let $M$ be any smooth surface embedded in $\mathbb{R}^3$. 
We consider a family of scalar functions defined on the surface $M$ and depending on a set of parameters $\chi$, i.e. formally defined as:
\begin{equation}
D_M ( \cdot, \chi): p \in M \subset \mathbb{R}^3 \longrightarrow D_{M}(p,\chi)  \in \mathbb{R}
\end{equation}

In this very general definition, $\chi$ corresponds to the set of parameters of the algorithm used to compute $D_M$ and allowing to adapt the error of the estimation.
Note that some of these parameters have to be adapted in particular depending on the size of the surface, i.e. relatively to the spatial discretisation of the surface (number of vertices in the corresponding triangular mesh) and/or of the ambient space (size of voxels).
Taking as an example the sulcal depth, the collection of parametric functions $\chi \rightarrow D_M(\cdot,\chi)$ represents all the possible settings to compute an estimation of the sulcal depth on $M$.
In the case of a method based on the distance to the convex hull, $\chi$ consists of all possible settings of the algorithm used to compute the convex hull, i.e. the size of structuring elements for a morphological closing of the volumetric segmentation of the brain, which has a direct influence on the estimated depth even if this influence is difficult to quantify.
In the case of the SULC method (described in section \ref{sec:SOTA} above), $\chi$ consists of all possible settings of the parameters from eq.\ref{eq:sulc} that controls for the inflation of the surface.

We define a scaled version of $M$ by applying a \textbf{scaling} of magnitude $s>0$, i.e. by multiplying the coordinates of $M$ by a the factor $s$, and we note this scaled surface $sM$.
In particular lengths have been multiplied by $s>0$.
\begin{definition}
We state that a family of functions $D_M$ defined on $M$ and depending on parameters $\chi$ is \textbf{scale invariant} if, for any scaling factor $s>0$ : 
\begin{equation}
    D_{sM}(\cdot,\chi) = D_{M}(\cdot,\chi)
\end{equation}
\end{definition}
Which means that the estimated value at each point $p$ is identical on $M$ and $sM$, for any setting of the parameters $\chi$.

Note that the scale invariance from Definition 1 is not very relevant in applications to real brains.
Indeed, as stated in the section \ref{intro}, variations between any two brains in global brain size are systematically associated with variations in folding magnitude.
When considering two cortical surfaces from different individuals, variations in brain size cannot be approximated by such simple \textbf{scaling} since brain volume and surface are related by an  allometry.
Even when considering the cortical surfaces from a single individual acquired at two different developmental stages, brain maturation impacts simultaneously both the scaling and the folding.
As a consequence, in the vast majority of applications, the experimenter aims at comparing the functions computed on $M_1$ and $M_2$, knowing that $M_2 \neq sM_1$.

We thus need to introduce a second definition:

\begin{definition}
We state that a family of functions $D_M$ defined on $M$ and depending on parameters $\chi$ is \textbf{scale controlled} if one can find $f(s,\chi)$ and $d(s)$ such as:

\begin{equation}
\forall p \in M:
\underbrace{\frac{1}{d(s)}}_{normalisation}D_{sM}(p,\underbrace{f(s,\chi)}_{adaptation}) = D_{M}(p,\chi)
\label{eq:norm_adapt}
\end{equation}
By construction it imposes $d(1)=1$ and $f(1,\chi)=\chi$.
\end{definition}

This equation makes a formal link between the estimations computed on the two surfaces $M$ and $sM$:
For any method $D_M$ satisfying this equation, the estimation $D_{sM}$ on a scaled surface is related to the estimation obtained on the original surface through two factors: 
\begin{itemize}
    \item a multiplicative function of $d(s)$ called \textbf{normalisation} that corresponds to normalising the range of the values covered across the points of each surface;
    \item an \textbf{adaptation} function $f$ that corresponds to the adaptation of the parameters
\end{itemize}
In other terms, this definition allows us to decompose the influence of the scaling in two factors: \textbf{normalisation} of the range of values and \textbf{adaptation} of the parameters.

\begin{definition}
We consider $M_1$, $M_2$ and parameters $\chi$. We define $s=L_2/L_1$ where $L_i$ is the characteristic length of $M_i$.
If $D$ is scale controlled we call $D^*$ the function satisfying:
\begin{equation}
    D^*_{M_2}(\cdot,\chi) = \frac{1}{d(s)} D_{M_2}(\cdot,f(s,\chi))
    \label{eq:Dstar}
\end{equation}
\label{def:3}
\end{definition}
In practice, we use $L_i=V_{M_i}^{1/3}$, i.e we take as the characteristic length of a surface the cubic root of its volume, and we face two possible situations:
\begin{itemize}
    \item If $M_2$ is a scaled version of the reference surface $M_1$, then 
    \begin{eqnarray*}
    D^*_{M_2}(\cdot,\chi) & = &   \frac{1}{d(s)}D_{M_2}(\cdot,f(s,\chi)) \ from\ equation\ Eq.\ref{eq:Dstar})\\
        & = & \frac{1}{d(s)}D_{sM_1}(\cdot,f(s,\chi)) since\ M_2=sM_1\\
        & = & D_{M_1}(\cdot,\chi) \ from\ equation\ Eq.\ref{eq:norm_adapt})\\
        & = & D^*_{M_1}(\cdot,\chi) \ from\ equation\ Eq.\ref{eq:Dstar}\ with\ s=1
    \end{eqnarray*}
    Thus $D^*_{M_2}$
    will be identically equal to $D^*_{M_1}$, which means that $D^*$ is scale invariant.
    In other words, in the ideal case of a perfect scaling between $M_2$ and $M_1$, a scale invariant can be defined on the basis of a scale controlled function by using Def \ref{def:3}.
    \item Now in the situation where $M_2$ is not a scaled version of $M_1$, we can nevertheless consider $s$ as an \textbf{empirical} scaling factor and quantify the error of $D^*$ to the perfect scaling ideal case. For that, we can consider that $D^*_{M_2}$ and $D^*_{sM_1}$ come from two random variables $X$ and $Y$ with probability distribution $p_X$ and $p_Y$.  Then considering any distance $Dist$ between probability distributions, the error will be:
    \begin{equation}
    \label{eq:wasserstein}
       \epsilon_{{M_1}, {M_2}}(\chi) = Dist(p_X,p_Y)
    \end{equation}
    In practice, we have used the 1D Wasserstein distance (also known as Earth Mover distance).
    This equation is a way to quantify the deviation to the scale controlled case. If $M_2=sM_1$, then $X$ and $Y$ corresponds to the same distribution and we have $\epsilon_{{M_1}, {M_2}}=0$
   
\end{itemize}

\subsection{Sulcal Depth with formal control on the scale of the surface}
\label{sec:DPF_star}

We now consider the application of the formalism provided above to a specific type of functions: sulcal depth.
In the present work, we consider the \textit{DPF} (introduced in Sec\ref{sec:SOTA}) as our method of interest. 
By definition through Eq.\ref{eq:DPF}, the $DPF$ is a family of functions with only one parameter $\alpha$. In the following we will denote it $D_M(\cdot,\alpha)$.

\begin{theorem}
The $DPF$ is scale-controlled and we have the explicit formula
\begin{align}
\label{eq:theorem1}
    f(s,\alpha) = s^{-2}\alpha \\
    d(s) = s
\end{align}
\end{theorem}

\begin{proof}
    We denote $M_2$ the surface obtained by applying a scaling of parameter $s$ to the surface $M_1$, i.e. $M_2=sM_1$. 

    We recall the intrinsic property of the Laplace-Beltrami operator and mean curvature:
\begin{equation} \label{eq:dedimensionalisation}
    \Delta_{M_2} = s^{-2} \Delta_{M_1} \,\
    K_{M_2} = s^{-1} K_{M_1} 
\end{equation}
    
    Equation \ref{eq:DPF} on $M_2$ with a parameter $\alpha_2$ can be transformed into an equation on $M_1$:

\begin{align*}
    & -\Delta_{M_2} D_{M_2} + \alpha_2 D_{M_2} = K_{M_2} \\
    \Leftrightarrow &-s^{-2}\Delta_{M_1}D_{M_2} + \alpha_2  D_{M_2} = s^{-1}K_{M_1} \\
    \Leftrightarrow &-\Delta_{M_1}D_{M_2} + \alpha_2 s^{2}  D_{M_2} = sK_{M_1} 
\end{align*}

Calling $\alpha = \alpha_2 s^2$, the previous equation can be written as

\begin{equation}
    -\Delta_{M_1}\big(s^{-1} D_{M_2}\big) + \alpha  \big(s^{-1} D_{M_2}\big) = K_{M_1} 
\end{equation}

The previous equation tells us that $s^{-1}D_{M_2}(\cdot,\alpha_2)$ is a solution of the Poisson equation on $M_1$ with parameter $\alpha$.
Since Poisson equation admits a unique solution, we conclude that : 
\begin{equation}
    s^{-1}D_{M_2}(\cdot,s^{-2}\alpha) = D_{M_1}(\cdot,\alpha)
\end{equation}
Eq. \ref{eq:norm_adapt} is satisfied and we get that $D$ is scale controlled.
\end{proof}

We can now apply Eq.\ref{eq:Dstar} to define the scale-controlled version of the $DPF$ that we call $DPF^*$.
So $DPF^*$ corresponds to a normalization of $DPF$ with respect to the variation in global scale between $M_1$ and $M_2$, and by computing $\epsilon_{{M_1}, {M_2}}(\alpha)$, we can quantify the potential deviation from the ideal case where   $M_2=sM_1$.

It is also useful to introduce $DPF^*_{abs}$ an absolute version of $DPF^*$ whose physical unit is a length: 
\begin{equation} \label{eq:DPF_abs}
DPF^*_{abs} = L \times DPF^*
\end{equation}
where $L$ is the characteristic length of the surface. 

\subsection{Adaptation of the scale of interest through $\alpha$ } 
\label{sec:alpha_setting}

Now let us focus on the setting of the only one parameter of \textit{DPF} (Eq.\ref{eq:DPF}), and thus of $DPF^*$ : $\alpha$.
As mentioned in \cite{boucher_depth_2009}, decomposing the DPF into the eigenfunctions basis of Laplace-Beltrami operator (Fourier modes) allows to better interpret the influence of $\alpha$. The eigenpairs $(\lambda_i,\Phi_i)$ are solutions of
\begin{equation}
\Delta \Phi_i = - \lambda_i \Phi_i
\end{equation}
where the $\lambda_i$  are positive since the operator is negative.
By expressing the mean curvature and the depth potential in this basis:
\begin{equation}
    D_{\alpha} = \sum_iD_i\Phi_i ,\, K = \sum_i K_i \Phi_i
\end{equation}
and injecting them in in (Eq.\ref{eq:DPF}) 
we obtain the following transfer function: 
\begin{equation} \label{eq:decomposition}
    D_i =  \frac{1}{\alpha + \lambda_i} K_i
\end{equation}

This formal analysis allows us to interpret the $DPF^{\star}$ as a low-pass filter on the curvature, the level of which is controlled by $\alpha$. 
Decreasing $\alpha$ increases the influence of the low frequency components of the curvature compared to high frequency components, and vice versa.
The low frequency components correspond to geometric information relative to the global shape of the brain, with large concave regions at the scale of a lobe (such as the prefrontal frontal and occipital regions).
Reducing the high frequency components reduces the influence of local folds and noise within the sulci. 
We refer to \cite{germanaud_larger_2012} for further interpretation of $\lambda_i$ in terms of wavelengths of an eigenfunction $\Phi_i$. 

The interpretation of the DPF* as a low-pass filter allows us to complement this exploration of the influence of $\alpha$ by analyzing its impulse response. 
The impulse response of the DPF is given by the Green's function $G^{\alpha}_p(\cdot)$ such that : 

\begin{equation}\label{eq:green_DPF}
    (-\Delta + \alpha I )G^{\alpha}_p(\cdot) = \delta_p(\cdot)
\end{equation}

where $\delta_p(\cdot)$ is the Dirac function in $p \in M$. 
We can thus express the $DPF$ as a convolution of the curvature with the Green's function: 

\begin{equation}
D_\alpha(p) = (K_{mean} \ast G^\alpha_p)(\cdot) 
\end{equation}

Note that Green's function is not identical at every point of the folded surface $M$ but depends on the local geometry of the surface.
Nevertheless, the impulse response can be interpreted in terms of the size of the neighbourhood which influences the estimation of the sulcal depth estimated at the location $p$.
We can thus interpret the value of $\alpha$ in terms of a \textit{scale of interest}: low (resp. high) values will correspond to large (resp. small) scales of interest.

\begin{figure}[h!]
 \centering
\includegraphics[width=0.7\linewidth]{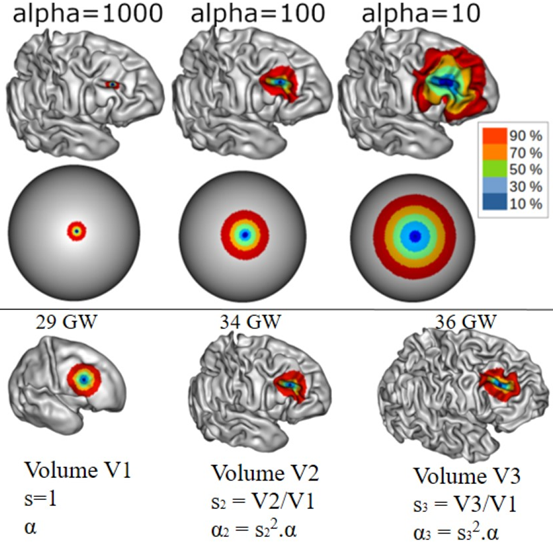}
\caption{Illustration of the dissociation between the value of the parameter $\alpha$ and the size of the surface.
Top: for both DPF and DPF*, larger values for $\alpha$ correspond to smaller scale of interest on a given surface. 
We also show the spherical representations to better visualise the variations in the scale of interest.
Bottom: thanks to the adaptation function $f$ in DPF*, once the scale of interest has been fixed for a given surface through the value of $\alpha$, it is automatically adapted according to the global size of the other surfaces.
}
\label{fig:alpha} 
\end{figure}

Note that since DPF* is scale-controlled, once we fix $\alpha$ for a given surface, this setting will be adapted to any other surface through the adaptation function $f$ from Eq.\ref{eq:theorem1}. 
In other words, with the DPF*, we can now dissociate the influence of the parameter $\alpha$ from the influence of the global size of the brain.
We illustrate this dissociation in Fig.\ref{fig:alpha}.
We pursue the description of the setting of $\alpha$ on real data in the next section. 

\section{Materials and Experiments}

We evaluate our method using quantitative and qualitative comparison with the closest related approach from the literature, the method SULC that was described in Sc.\ref{sec:SOTA}.
In this work, we compute SULC using the deform-mesh process available in the bioMedIA mirtk pipeline \footnote{https://github.com/BioMedIA/MIRTK}. 
We confirmed that the estimation is very close to the one from Freesurfer by computing the correlation between the two implementations for several brains and observed a correlation coefficient systematically superior to .99.
We present five complementary experiments:
\begin{itemize}
    \item In the first experiment, we explore the influence of the only parameter of our method ($\alpha$ from Eq.\ref{eq:Dstar}). 
This experiment allows us assess the influence of this parameter, and to identify a quasi-optimal setting.
\item In the second experiment, we provide empirical evidence that our method is scale-controlled, and show quantitatively that it is not the case for SULC. 
\item In the third experiment, we demonstrate the practical relevance of our approach by analyzing the influence of global brain size on the estimated sulcal depth across a large population of 1987 subjects.
\item In the fourth experiment, we compare our method and SULC on the surfacic atlas from \cite{bozek_construction_2018} that is representative of the evolution of brain folding with age during normal early development.
\item The fifth experiment consists in assessing the robustness of our method with respect to the spatial resolution of the mesh.
\end{itemize}


Of note, the combination of these experiments also constitutes a comprehensive experimental framework allowing to evaluate and compare all sorts of sulcal depth estimation methods.
We openly share the materials and code in order to ease the reproduction of our work, and to enable the use of our evaluation framework in future works introducing potential improvement to sulcal depth estimation techniques. 

\subsection*{Data}
In line with the objectives of our work, we aggregated a collection of cortical surfaces showing strong variations in brain size and magnitude of cortical folding.
To do so, we selected individual data from publicly available datasets: 

1) The third release of the dHCP dataset (http://www.developingconnectome.org \cite{edwards_developing_2022}) consists of 887 MRI sessions from 783 newborn babies covering the ages from 26 to 45 weeks post-conception.
We used the cortical surfaces ('.white' files) provided by the consortium. The segmentation and surface extraction tools are detailed in \cite{makropoulos_developing_2018}. The quality of the surfaces and corresponding segmentation were checked before the release to confirm the absence of obvious errors.

2) The cortical surface atlas was released with the corresponding article \cite{bozek_construction_2018} and is available online \footnote{https://brain-development.org/brain-atlases/atlases-from-the-dhcp-project/cortical-surface-atlas-bozek/}.
This atlas consists of age-dependant surface meshes spanning 28-44 weeks of gestation, constructed using adaptive kernel regression applied to surface data from the dHCP dataset (described above).

3) The KKI (Kennedy Krieger Institute, \cite{landman_multi-parametric_2011}) dataset consists of high quality MRI data from 21 healthy adults.
The data collection and quality assessment has been described in \cite{landman_multi-parametric_2011}. 
We downloaded the raw MRI data and used the recon-all pipeline from freesurfer 6.0 to extract the cortical surfaces as described in \cite{dale_cortical_1999}. 
We checked visually the extracted surface for all the 21 subjects to avoid any potential influence of inaccuracy in the surface extraction process.

4) The Human Connectome Project (HCP, \cite{van_essen_human_2012}, S1200 release) dataset consists of the high quality data from 1113 healthy young adults. 
Structural images were processed using the HCP structural processing pipeline, which has been described in \cite{glasser_minimal_2013}.
We used the preprocessed data (tissue segmentation, cortical surface extraction and registration) released by the consortium.

Each study was approved by their respective research ethics committee.
As detailed below, we selected different subsets of these datasets depending on the objective of each experiment.

\subsection{Experiment \#1 : setting of the parameter alpha of the $DPF^*$ }
\label{sec:desc_expe1}
The aim of this experiment is to assess the influence of the $\alpha$ parameter of $DPF^*$ and to define its most efficient setting.
To do so, we computed the DPF* with various values of $\alpha$ ranging from 0 to 2000 on the set of selected brains described below.
We also computed the SULC and DPF estimations, as well as the mean curvature (denoted as CURV) using the method described in \cite{rusinkiewicz_estimating_2004} in order to quantitatively compare the methods on the same set of surfaces.

We carefully selected a set of 16 individual surfaces as follows:
First we excluded all the subjects from the perinatal dataset dHCP for which any kind of incidental finding was detected by a neuroradiologist (radiology score superior to one), and we selected 13 subjects. 
In order to further extend the range of size in this dataset, we selected 3 adult subjects from KKI: the one having the largest, the lowest and an intermediate surface area.
Using this procedure, we obtained our dataset \#1 consisting of 13 perinatal subjects with an age range of [29 GW ,44 GW] and a range of volume of [50000 $mm^3$, 225000 $mm^3$];
and of 3 young adult subjects with a volume of respectively $324891 mm^3$, $468156 mm^3$, $608843 mm^3$.

For each of these 16 individual surfaces, we traced manually a set of specific geometrical landmarks, allowing to quantitatively assess three complementary desirable features relative to the anatomical relevance of a given sulcal depth estimation method.
For each feature, we designed the corresponding metric to be computed based on specific anatomical landmarks as described below. We provide in Appendix \ref{app:protocol} an extensive description of the protocol we designed to enforce a reproducible manual tracing of these anatomical landmarks. We also share the traced landmarks \footnote{link textures} allowing to evaluate on the same data the future sulcal depth estimation methods proposed by other researchers.

\begin{itemize}
    \item \emph{Feature \#1}: The sulcal depth should have a high and relatively stable value (low standard deviation) along the crests of major gyri, since all these gyral crests are located very close to the convex hull of the brain. \\
    \emph{Landmark \#1}: We manually traced gyral crests as the lines surrounding sulcal basins, at the crown of gyri, close to the convex hull of the brain. We note the set of gyral crests from a given surface as $\mathcal{C}$.\\
    \emph{Metric \#1}: $StdCrest$ is computed as the standard deviation of depth along gyral crests, with a normalisation to compensate for variations in brain size across subjects:
    \begin{equation}
        StdCrest(D)= \frac{std(D\rvert_{\mathcal{C}})}{IPR(D)}
    \end{equation}
    where : 
    \begin{equation}
        IPR(D) = \textit{percentile}(D,95) - \textit{percentile}(D,5)
    \end{equation}
    where $std(D\rvert_{\mathcal{C}})$ is the standard deviation of $D$ across vertices located on the gyral crests.
    Since we want to evaluate the variation in depth values on the crests in relation to the variation in depth across all the vertices of $M$ (and not only on $\mathcal{C}$), we normalise according to $IPR(D)$.
    $IPR(.)$ stands for inter-percentile range with $percentile(D,x)$ being the $x$th percentile of the depth distribution across all the vertices of $M$.
    $IPR(.)$ is a robust estimation of the spread of a distribution \cite{ross_introductory_2017}.
    Using this metric, an ideal sulcal depth map would have a $StdCrest$ value close to 0. 

    \item \emph{Feature \#2}: The higher values of sulcal depth should align with gyral crests, while the lower values should align with sulcal fundi.\\
    \emph{Landmark \#2}: The fundus lines are traced along the bottom of sulcal basins.
    We denote the set of sulcal fundi as $\mathcal{F}$.\\
    \emph{Metric \#2}: $Sep$ quantifies the distance separating the median of depth values corresponding to gyral crests from the median of depth values corresponding to sulcal fundi: 
    \begin{equation}
        Sep(D)= \frac{1}{IPR(D)}(median(D\rvert_{\mathcal{C}}) - median(D\rvert_{\mathcal{F}})
    \end{equation}
    $Sep$ varies theoretically between 0 and 1. A value of 1 corresponds to the ideal case where the crests have a constant and maximal depth value and the fundi a minimal and constant value. This is not realistic as the fundi are not expected to have a constant depth value. We also expect that the range of values on the fundi is larger for adult brains compared to less folded baby brains due to the appearance of secondary and tertiary sulci. In any case, a low $Sep$ (below 0.5) would be indicative of inconsistent sulcal depth values, with relatively high values even in the fundi and/or relatively low values on the gyral crests.

    \item \emph{Feature \#3}: The gradient of the sulcal depth should align with the direction of the shortest path between sulcal fundi and corresponding gyral crests.\\
    \emph{Landmark \#3}: The shortest paths between fundi and gyral crests are computed automatically by connecting a vertex from the crest to the closest vertex of a fundi through the path with lowest geodesic length. We denote $\mathcal{W}$ the set of shortest paths computed for a given mesh $M$. \\
    \emph{Metric \#3}: $Dev$ quantifies the angular deviation between the gradient of the sulcal depth map and the direction of the shortest paths:
    \begin{equation}
        Dev(D) = \frac{1}{\# \mathcal{W}} \sum_{p \in \mathcal{W}} angle(p)
    \end{equation}
    Where $angle(p)$ is the angle between the gradient of the sulcal depth map and the direction along the shortest path computed at a point $p\in W$.

    $Dev(.)$ theoretically varies between 0 and 180 degrees.
    An angular deviation close to 90 degrees is indicative of an undesirable depth estimation such as a minimum located on the sulcal wall or a pinch along the sulcal wall interpreted as a gyral crest. Values greater than 90 degrees are suspicious and may occur only in the cases of complex geometrical configurations e.g. due to operculation. 

\end{itemize}

We report in Fig.\ref{fig:expe1} and corresponding Sec\ref{sec:res_expe1} the distribution of these three measures across all subjects as violin plots.
Independently for each measure, we then identified the best value for $\alpha$ based on the median across all subjects (lowest median for  $StdCrest$ and $Dev$, highest for $Sep$). 
For each measure, we then performed a non-parametric Wilcoxon signed-rank test between each distribution obtained for the different values of $\alpha$ and the distribution corresponding to the best $\alpha$, allowing us to define a range of values on which the measures are statistically equivalent.

\subsection{Experiment \#2 : Empirical confirmation of the invariance to scaling}
\label{sec:desc_expe2}

In this experiment, we investigate the influence of variations in global scaling of a surface on the sulcal depth estimated at each location of the mesh using our method $DPF^*$ and SULC.



To do so, we defined the dataset \#2 consisting of a set of 5 synthetic surfaces generated by applying several global scaling to a real surface. 
We selected a mesh from a young brain (29GW) from the dHCP and we applied a global scaling of magnitude 2, 3, 4, and 5 that resulted in 5 surfaces (incl. the scale 1) with exactly the same shape but different global sizes : $65000 mm^3$, $520000 mm^3$, $1755000 mm^3$, $4160000 mm^3$, and $8125000 mm^3$. 

We then computed the SULC and $DPF^*$ on each of these 5 surfaces. 
In order to characterise the influence of a given scaling on the estimated sulcal depth, we computed the linear regression across all the vertices of the mesh between the depth estimated on the scaled surface and the depth estimated on the original surface (scale=1).
If the method is scale invariant, the correlation coefficient should be equal to 1 and the slope of the regression line should also be equal to 1.
If the global scaling affects the estimated sulcal depth, then the regression line might deviate from $y=x$, and/or the correlation coefficient might be lower than 1.


\subsection{Experiment \#3 : Application to a large dataset}
\label{sec:desc_expe3}

In this  experiment, we compare our method with SULC in the context of a concrete application to a large set of real brains.
The dataset \#3 consists of the cortical meshes from 887 scanning sessions from the dHCP neonatal dataset, and 1108 subjects from the HCP dataset.
The computation of sulcal depth failed for 8 subjects from dHCP and 5 subjects from HCP likely due to topological defects in the cortical mesh, resulting in a total of 1987 surfaces with total volume ranging from 33469$mm^3$ to 141421$mm^3$ for the dHCP dataset and from 182926$mm^3$ to 383241$mm^3$ for the HCP dataset. 

We computed SULC, the $DPF^*$ and $DPF^*_{abs}$ for each of these surfaces and show the distributions across the vertices of the corresponding mesh of the resulting sulcal depth measures as a function of the surface area.
In addition, we computed the Wasserstein distance as proposed in Eq.\ref{eq:wasserstein}, between every pair of individuals, resulting in a matrix of Wasserstein distances for each method. As explained in Sec.\ref{sec:scale_control}, this allows us to quantify the deviation to the scale controlled case for the $DPF^*$, and to quantify the influence of global brain size on the estimated sulcal depth measures.
We then assessed whether the Wasserstein distances computed for larger brains were statistically different from the distances computed for smaller brains.
To do so, we selected 8 subgroups of 200 subjects corresponding to subpopulations of different brain size, ordered from smaller brains to larger brains.
This approach corresponds to a sliding window analysis of the patterns observed in the distance matrices.
We then set as a reference subgroup the 200 subjects with larger brain size and assessed statistically putative differences in the distribution of Wasserstein distances between this subpopulation and the other subgroups using the Kolmogorov-Smirnov statistic for two samples.
The resulting statistic, with values between $0$ and $1$, indicates how similar are the distributions from the other subgroups with respect to the group with larger brains.
In other words this statistic is a quantitative measure of the influence of the variations in brain size on the estimated sulcal depth maps:
A low statistic value would mean that the sulcal depth maps estimated for the two subgroups are similar despite variations in global brain size.  
A high statistic value would correspond to the case where the distributions are very different.

\subsection{Experiment \#4 : Quantifying brain folding dynamics}
\label{sec:desc_expe4}

In this experiment, we compare SULC and $DPF^*_{abs}$ at a local scale by assessing the dynamics of the formation of the main cortical suci.
To do so, we applied the different estimation techniques to the spatio-temporal surfacic atlas computed from the dHCP dataset as described in \cite{bozek_construction_2018}, and compared the depth values in 23 sulcal regions of interest distributed across the entire cortex.
This atlas is representative of the evolution of the cortical surface between 28 and 44 weeks post-conception and thus shows marked changes in both the global size and in the magnitude of folding.
For each of the 16 time steps available in the atlas, we applied SULC and $DPF^*_{abs}$ independently.
The delineation of the sulcal regions was done by semi-automated tracing on the oldest atlas (44 weeks) as follows:
We first extracted all the sulcal basins defined as the regions with negative values in the smoothed mean curvature map.
We then manually labeled a set of sulcal basins according to the nomenclature provided in \cite{de_vareilles_development_2023}, aiming at covering most of the cortex, with sulci expected to appear at different gestational ages.
We did not label the sulcal basins corresponding to very small sulci for which the correspondence across the different time steps of the atlas would not be anatomically accurate.
We obtained 23 sulcal regions corresponding to sulci appearing across the entire age range covered by the atlas.
Since the atlas was designed using surface-based registration ensuring correspondence in both space and time (a given vertex on the mesh can be mapped across the different time steps, see \cite{bozek_construction_2018} for details), the sulcal regions defined in the last time step (44 weeks) can be projected onto earlier time steps, and the anatomical relevance of the definition of the sulcal regions can be assessed visually, as shown on Fig.\ref{fig:atlas_sulci}.
This approach allowed us to delineate the 23 sulcal regions for all the time steps of the atlas, and we were able to track the evolution with age of the estimated depth from the different techniques.

\subsection{Experiment \#5 : Assessment of the potential influence of the spatial resolution of the mesh}
\label{sec:desc_expe5}

In this experiment, we assess the potential impact on the $DPF^*$ of variations in the spatial resolution of the mesh, i.e. in the number of vertices and triangles in the mesh.
To do so, we use the 16 cortical surfaces of experiment \#1 that cover a large range of levels of cortical folding, and apply a the down sampling scheme \cite{garland1997surface} as implemented in the PyMeshlab package \cite{pymeshlab}. For each surface we obtain 3 levels of lower resolution by keeping 25, 50, 75 \% of the vertices. We then compute the $DPF^*$ for all the surfaces and project the depth map computed on down-sampled meshes onto the corresponding original, high-resolution, mesh by using a nearest neighbor interpolation in 3D. 
We finally compute the normalized root mean squared error (NRMSE) expressed as a percentage of the value from the original mesh:
\begin{equation}
    NRMSE(DPF_{orig}, DPF_{proj})=100 \times \frac{\sqrt{\frac{1}{N_{low}} \sum_{i=1}^{N_{low}} (DPF_{orig}(i) - DPF_{proj}(i))^2}}{\sqrt{\frac{1}{N_{low}} \sum_{i=1}^{N_{low}} ( DPF_{proj}(i))^2}} 
\end{equation}

\section{Results}

\subsection{Result from experiment \#1: setting of parameter $\alpha$}
\label{sec:res_expe1}

\begin{figure}[h!]
    \centering
    \includegraphics[width=0.72\linewidth]{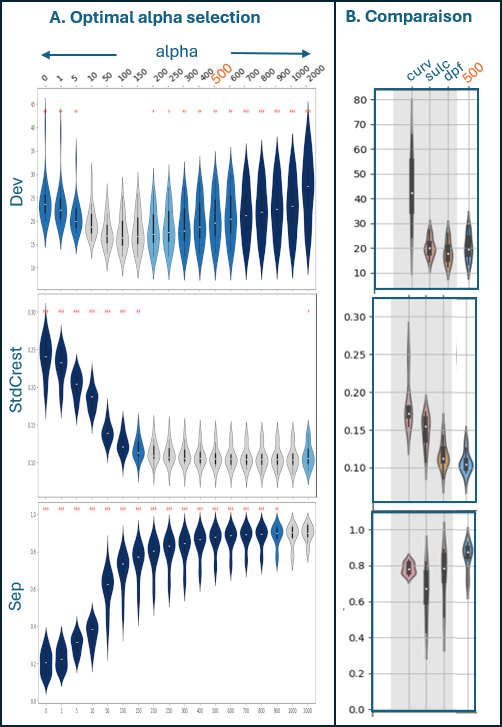}
\caption{Results from Expe \#1.
A: From top to bottom, we show the distributions of the three measures $Dev$, $StdCrest$, and $Sep$ across all the brains from the dataset \#1 as violin plots, for the different values of $\alpha$. 
We highlight in gray the range of values for which the distributions are statistically equivalent to the one obtained using the best value for $\alpha$ (Wilcoxon-test with p-value$>$.05).
B: For each measure, we show the distributions obtained for the method SULC, the mean curvature and the classical $DPF$ from \cite{auzias_deep_2015}, for comparison.
}
\label{fig:expe1} 
\end{figure}

The results of Expe \#1 are shown in Fig.\ref{fig:expe1}
On this figure, we report the distributions from the three measures detailed in Sec.\ref{sec:desc_expe1} ($Dev$, $StdCrest$, and $Sep$) computed across the dataset \#1, for different values of the $\alpha$ parameter.

From the plot of the angular deviation $Dev$ in Fig.\ref{fig:alpha}.A, we observe that for $\alpha = 0$, the median of the distribution is around 25 degrees.
$Dev$ decreases to 16 degrees for $\alpha = 50$, and then increases to almost 30 degrees for $\alpha = 2000$. 
The variance of the distribution is also lower for $\alpha = 50$ than for lower and larger values of $\alpha$.
The statistical tests indicate that the distributions for $Dev$ are not statistically different for $\alpha\in [10, 150]$.

From the plot of the normalized standard deviation on the crests $StdCrest$, we observe a strong decrease with increasing $\alpha$, from .24 for $\alpha = 0$ to .12 for $\alpha = 150$.
The lowest median is obtained for $\alpha=400$, and the statistical tests show that the distributions are equivalent for $\alpha \in [200, 1000]$.

From the plot of the difference between the median of the crest and the median of the fundi $Sep$, we observe a progressive increase with increasing values of $\alpha$.
From a value of .2 for $\alpha = 0$, the median of $Sep$ increases up to .9 for $\alpha = 500$, with a highest median for $\alpha=2000$.
The statistical tests indicate that the distributions are equivalent for $\alpha$ values above 1000.

Consistently with our observations from Fig\ref{fig:alpha}, the spatial distributions of the estimated sulcal depth throughout the entire surfaces indicates that low values of $\alpha$ correspond to large scale of interest, leading to overestimated depth values in the fundi that are affected by the values obtained in highly concave regions such as the anterior temporal lobe, the occipital lobe and the prefrontal cortex. 
This induces the lower values for $Sep$ and higher value for $StdCrest$ when $\alpha$ is set to lower values.
On the opposite, for very high values of $\alpha$, the scale of interest becomes too small.
$Sep$ can be close to 1 and $StdCrest$ close to 0 as for the mean curvature, but the angular deviation $Dev$ increases, indicating that the influence of very small noisy bumps on the surface becomes too strong.  

When comparing with the SULC and classical $DPF$ methods in Fig.\ref{fig:alpha}.B, we observe that their angular deviation $Dev$ are also quite low compared to mean curvature, and not different from the best setting of $\alpha$ for $DPF^*$.
These distributions confirm the visual observation of the appropriate setting of the default parameters of SULC for brains having a global scale not too different from an adult.
Note however that for $StdCrest$ and $Sep$, the distributions from $DPF^*$ with the best $\alpha$ are statistically different from SULC, and from the classical $DPF$.

Overall, we observe that the influence of $\alpha$ on the estimated suclal depth from $DPF^*$
is continuous, since a small variation of $\alpha$ will have a very small influence.
In addition, $DPF^*$ shows better measures than both DPF and SULC for a large range of values of $\alpha$. 
Considering an optimal value $\alpha = 50$ for $Dev$, $\alpha=400$ for $StdCrest$, and the plateau after $\alpha=1000$ for $Sep$, we choose an intermediate value of $\alpha=500$ for the rest of the experiments. This choice was corroborated by visual inspections.

We provide in Appendix 6.6 a visual comparison between SULC and $DPF^*$ with  $\alpha=500$.

\subsection{Result from experiment \#2: invariance to scaling}

\begin{figure}[h!]
\includegraphics[width=\linewidth]{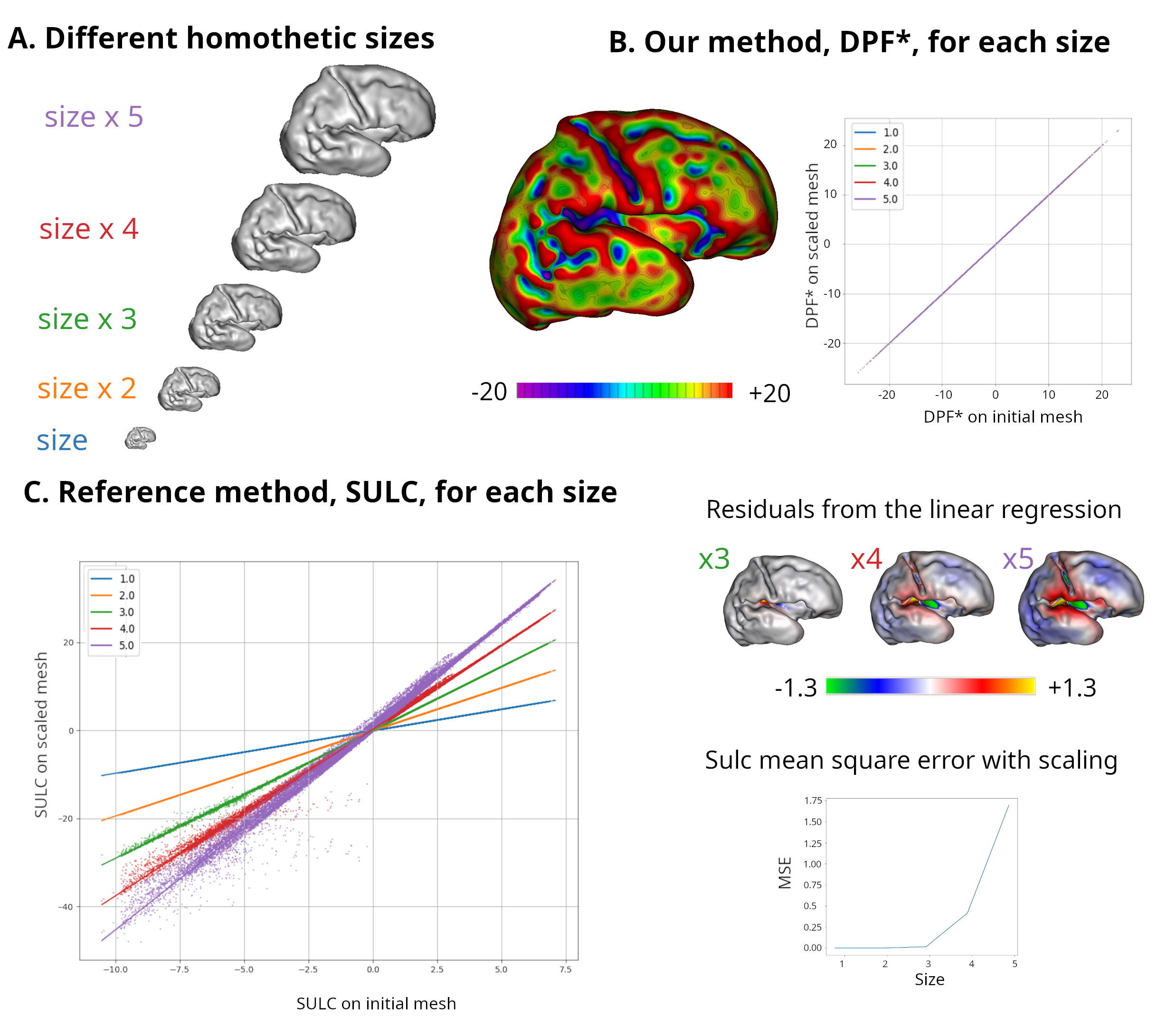}
\caption{Results from Expe \#2.
A. The scaled surfaces obtained as described in Sec.\ref{sec:desc_expe2}.
B. Depth maps generated using the $DPF^*$ method (with $\alpha=500$). The scale-invariance is confirmed empirically: the estimated depth values are not influenced by the scaling. The correlation is thus perfect for all surfaces and the regression lines are super-imposed.
C. Left: Depth estimations resulting from the SULC method applied to the scaled surfaces (y-axis) with respect to SULC on the initial mesh (x-axis). 
Both the slope and the correlation values depend on the scaling factor.
Right: Spatial distribution of residuals from the linear regression models derived from the left plot.
Consistently with the regression plot, the residuals increase with the scaling factor.
The regions corresponding to most concave and convex geometric configurations show higher magnitude of error.
}
\label{fig:exp_hom} 
\end{figure}

The results of Expe \#2 are shown in Fig.\ref{fig:exp_hom}.

First, this experiment allows us to confirm empirically that the $DPF^*$ is scale-invariant: $ D^*(M, s) = D^*(M,1)$ for $s = {2,3,4,5}$.
This experimental confirmation aligns with the theoretical formulation we introduced earlier.

In contrast, our examination of the SULC method revealed its sensitivity to the global scale of the mesh: if the mesh is two times larger, the depth estimated at a given location is roughly two times larger. 
The regression analyses indicate that a normalisation factor would compensate for most of the variations induced by the global scaling. 
However the spatial distribution of the residual errors (right panel of Fig.\ref{fig:exp_hom}.C) confirms that such a normalisation would not be sufficient to compensate for the influence of the global scaling factor on the depth.



\subsection{Result from experiment \#3: Application to a large dataset}

\begin{figure}[h!]
\includegraphics[width=\linewidth]{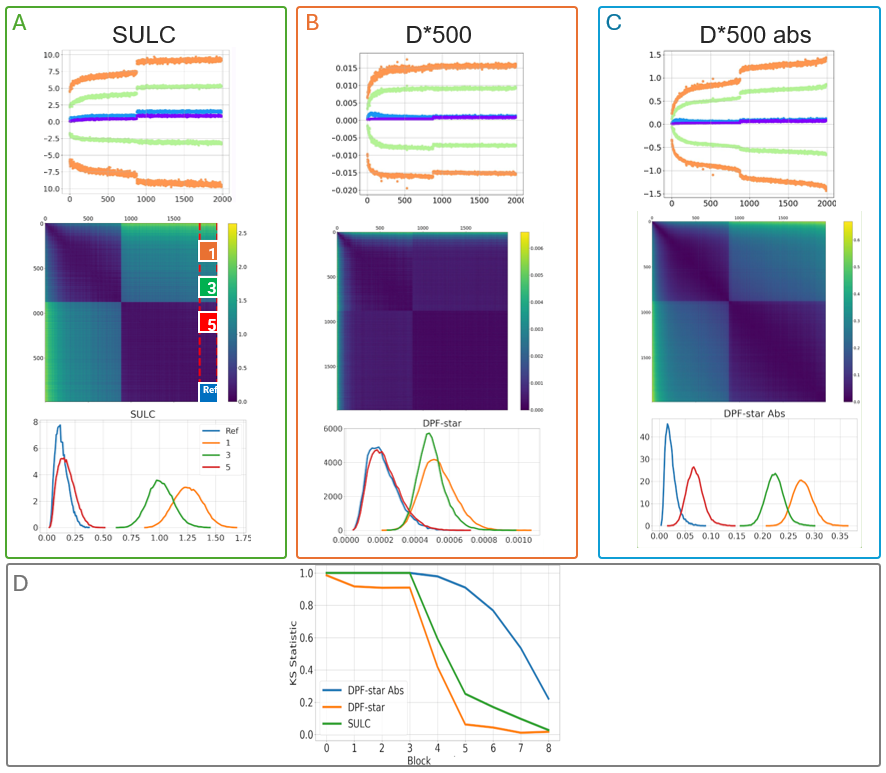}
 \caption{Analysis of the distribution of SULC, $DPF^*$ and $DPF^*_{abs}$ across the 879 surfaces from the dHCP newborns and 1108 surfaces from HCP young adults.
 From left to right, the three columns correspond resp. to SULC, $DPF^*$ and $DPF^*_{abs}$.
 On the top row, we show the distribution of each sulcal depth estimation for all subjects, ordered following their global size. For each method, we show the mean in purple, the median in blue, the 25th and 75th centiles in green, and the 5th and 95th centiles in orange.
 On the second row, we show the corresponding matrices of the Wasserstein distance between every pair of surfaces.
 On the third row, we show for each method the distributions of the Wasserstein distances within the four sub-populations corresponding to the blocks illustrated on the matrix for the method SULC.
 At the bottom, we show for each method the KS statistic comparing the distribution in the block corresponding to the subgroup of largest brains, to the other blocks corresponding to smaller brains. See the text for further description.
}
\label{fig:expe3} 
\end{figure}

The results of Expe \#3 are detailed in Fig.\ref{fig:expe3} and corresponding caption.
The distributions of the different sulcal depth estimation techniques (top row) are indicative of the influence of the global brain size.
For SULC, on the left, the impact of the increase in brain size from left to right is visible on all the centiles.
More specifically, the jump in the dispersion of the distributions (esp. visible on the 75th and 95th centiles) corresponds to the transition from the 879 newborns from the dHCP to the adults from the HCP.
On the dHCP dataset, the dispersion progressively increases with brain size, while on the HCP dataset, the distributions are very similar across individuals despite variations in global brain size (52\% of variation between the volumes of the smallest and largest adults from HCP).

The distributions of the $DPF^*$ method, at the center, illustrate the efficiency of our formal normalisation.
The influence of brain size on the centiles strongly reduced compared to SULC.
More specifically, the normalisation compensates for variations in brain size across the dHCP or HCP datasets, since the transition is barely visible in the distributions.
We also observe that the influence of brain size on the distributions is still important for very small brains.
Indeed, the variations in brain size are huge for the first 200 subjects that correspond to younger babies with volume values ranging from 7584$mm^3$ to 33708$mm^3$.

For the $DPF^*_{abs}$, on the right, we see that by multiplying the $DPF^*$ by the characteristic length as in Eq.\ref{eq:DPF_abs}, we obtain an absolute estimation of the sulcal depth with a clear influence of the global size of the mesh, as expected.
The increase in the dispersion of the distributions with brain size is clearly visible and progressive.
Of note, the influence of brain size is stronger than for SULC on the adult subjects from the HCP dataset.
The $DPF^*_{abs}$ shows a higher sensitivity than SULC to the 52\% variation in brain volume across adults.

On the second row, the Wasserstein distance matrices computed for each method independently show patterns that are consistent with the observations from the distributions. 
For SULC, we observe very low Wasserstein distance values across adult individuals from the HCP, and high values when the distance is computed between subjects from the HCP and dHCP datasets.
For $DPF^*$, the distance values are very similar between the larger brains from the dHCP and the adults from HCP, which confirms that the influence of global brain size is compensated.
Consistently with the distributions, the distance values are higher for the 200 younger, smaller, less folded brains.
The pattern for $DPF^*_{abs}$ is very similar to SULC, with a slightly more progressive evolution between dHCP and HCP individuals.

On the third row, the distributions of Wasserstein distances across the subjects within the reference group are the leftmost (lower distance values), for each of the three methods. The Wasserstein distances computed between the reference group and other subpopulations show a progressive convergence toward the reference distribution, when the brain size inside the subpopulations increase. This effect is already present for small brains for the $DPF^*$ (subpopulations 1 and 3) but not for $DPF^*_{abs}$ and SULC.

At the bottom, the Kolmogorov-Smirnov statistic is used to compare quantitatively the distributions of Wasserstein distances across the four sub-populations shown on the third row.
We observe a transition between blocks 3 and 5, since the sub-populations switch from the babies from dHCP to the adults form HCP. 
Note that the K-S statistics for $DPF^*$ is always below the one of SULC, which confirms quantitatively the lower influence of brain size on $DPF^*$. 
For $DPF^*_{abs}$, we observe a much smoother curve illustrating the progressive influence of the global scale on the depth estimation, from babies to small adult brains, to larger adult brains.


\subsection{Result from Experiment \#4 : Quantifying brain folding dynamics}
\label{sec:res_expe4}

\begin{figure}[h!]
\includegraphics[width=\linewidth]{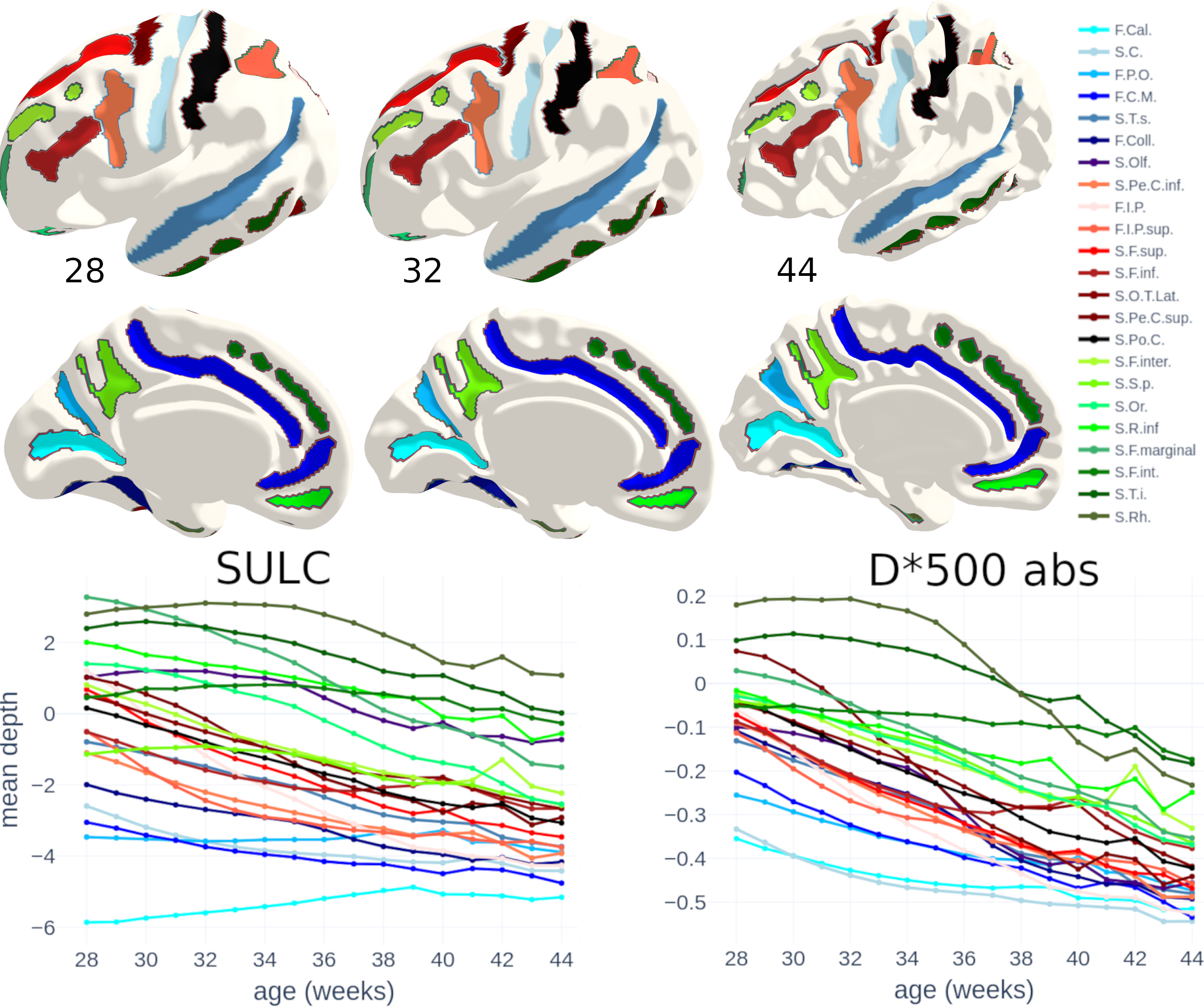}
\caption{Top: Labeled sulcal regions shown on three time steps of the spatio-temporal atlas.
The nomenclature of sulci and associated colors are shown on the right.
The colors are indicative of the expected timing of formation: sulci colored in levels of blue are expected to appear before the sulci in red that appear before the sulci in green.
Bottom: Curves showing the average sulcal depth estimated in the different sulcal regions for SULC (left) and $DPF_{abs}^*$ (Right).}
\label{fig:atlas_sulci} 
\end{figure}

We show in Fig. \ref{fig:atlas_sulci} the 23 sulcal regions on the surfacic atlas from the dHCP \cite{bozek_construction_2018} at 28, 32 and 44 weeks post conception, for the left hemisphere.
We provide in Appendix 6 the same illustration for the right hemisphere, as well as an illustration of the estimated suclal depth mapped onto the atlas. 
See \cite{de_vareilles_development_2023} for the complete description of the nomenclature used to label the sulci.
In the bottom panel, we show the average of the sulcal depth within each sulcal region as a function of the age for SULC on left and $DPF_{abs}^*$ on the right.
The color of each curve corresponds to the sulcus nomenclature shown at the top right of the figure.
From the literature of early brain development (see \cite{de_vareilles_development_2023} for a review of the timing of formation of the sulci), we expect to observe 1) a decrease in the metrics reflecting the depth for all sulci and 2) a specific order across sulci in terms of dynamics of the deepening (some sulci becoming deeper before others) and in terms of relative overall depth (some sulci being deeper than others at all ages).
The two observations are confirmed for $DPF_{abs}^*$ (on the right): according to \cite{de_vareilles_development_2023}, the sulci colored in levels of blue are expected to appear before and be deeper than the sulci in red, which are expected to appear before the sulci in green.
In addition, the decrease in depth metrics is clear for all sulci except for later forming sulci such as S.T.i and S.Rh for which the curve stays flat until 34-36 weeks and then the depth metric starts to decrease.
The expected ordering of sulci according to their overall depth is well respected, with blue curves corresponding to deeper sulci systematically located below the red and green curves.

Regarding the SULC method (on the left), this ordering according to the overall depth is relatively preserved, but with noticeable exceptions such as the Olfactory Sulcus (S.Olf, shown in dark blue), for which the curve is located within the group of green curves (shallow folds) and above the red curves that are supposed to be shallower than S.Olf.
In addition, we observe an unexpected reduction in depth with age (increase in corresponding curves) for the Calcarine Fissure (F.Cal, in cyan) and the Parieto-Occipital Fissure (F.P.O in light blue).
These two sulci are expected to be already present at 28 weeks, but we expect they become deeper on this period, since the global brain size increases.


In this experiment, we thus observe that the $DPF_{abs}^*$ allows us to characterize the dynamics of the formation of main cortical sulci. The biological interpretation is more consistent with the literature than with SULC.

\subsection{Result from experiment \#5 : Assessment of the potential influence of the spatial resolution of the mesh}

We report in Table \ref{tab:resolution} the comparison between the sulcal depth estimated on the resampled surfaces and the estimation from the original mesh. 
We report the mean and standard deviation of the NRMSE across the 16 subjects. 
We observe that the influence of the resolution is limited, with differences reaching only 15 \% for meshes that have been down-sampled by a factor 4.

\begin{table}[h]
    \centering
\begin{tabular}{|c|c|c|c|}
   Resolution & 75 \% & 50 \% & 25 \%   \\
    \hline
   NRMSE (\%$\pm$ std) & 3.1 $\pm$ 0.4 & 6.3 $\pm$ 0.6 & 15.4 $\pm$ 0.9  \\ 
\end{tabular}
    \caption{
Evolution of the mean and std of NRMSE with the spatial resolution}
    \label{tab:resolution}
\end{table}

In addition, we note that the computation time of $DPF*$ is less than 1 minute for a mesh of $~$60k vertices on a regular laptop.

\section{Discussion and Perspectives}

In this work, we proposed a formalization of the influence of global scaling on functions defined on surfaces, by introducing the properties of scale-invariance and scale-control. 
It allowed us to dissociate two factors: the normalisation of the amplitude of the function and the adaptation of the parameters of the function.
We then applied this formalism to the specific case of sulcal depth estimation by introducing the $DPF^*$, a scale-controlled version of the $DPF$, and the corresponding absolute measure $DPF^*_{abs}$. 

\subsection{The importance of global normalization}

The neuroimaging literature is increasingly invoking the sulcal depth as a relevant feature for characterizing brain morphology, in particular during development.
Sulcal depth has been used to describe the spatio-temporal evolution of brain folding in human fetuses and newborns \cite{yun_temporal_2020, xu_spatiotemporal_2022, lefevre_identification_2009}, and between birth and adulthood \cite{hill_similar_2010}. 
Recent studies suggest that sulcal depth in fetuses provides prognostic value of later cognitive development \cite{bartha-doering_fetal_2023, hahner_global_2019}, and is instrumental to detect neuro-developmental malformations \cite{im_atypical_2016, brun_localized_2016}.
Sulcal depth is also used to validate biomechanical models of gyrification \cite{wang_influence_2021, alenya_computational_2022}.
In all these exemplar studies, the use of our $DPF^*$ would enable better interpretation.

Indeed, when confronted to the issues related to comparing morphometric features between brains of different sizes, various strategies have been experimented. For instance, in \cite{yun_temporal_2020}, the authors compared cortical folding patterns across fetuses at different ages, and with a group of adults. Smoothed mean curvature was used as a proxy of sulcal depth in fetal brains in order to detect small and shallow folds, while the method in \cite{yun_automated_2013} was used in adults. The use of different methods at different ages limits the interpretation to large effects.
In \cite{meng_spatial_2014}, the authors decomposed the cortex in sulcal basins. To compensate for variations in brain size, the  parameters of their algorithm were adapted using linear regression models including as main factors surface area and maximum depth.
In these works, the normalisation is empirical, and the improvements compared to non-normalised estimation techniques is only qualitative. The $DPF^*$ proposed in this work enables to leverage these serious limitations.

\subsection{Literature on formal normalisation of surface features}

To our knowledge, the notion of scale invariance as a "desirable property of measures of brain shape" was first proposed in \cite{batchelor_measures_2002}. The authors illustrated the problem for curvature and various gyrification indices. Similarly in \cite{pienaar_methodology_2008} the authors documented the changes in growth and folding during the development, and their impact on curvature measures.
However these seminal works did not propose any implementable solution. 
A pragmatic and formal strategy called \textit{Nondimensionalisation} was proposed in \cite{rodriguez-carranza_framework_2008} and used in \cite{knutsen_spatial_2013}. The principle consists in dividing a folding measure by an estimate of global brain size in order to obtain a dimensionless quantity.
This global metric was denoted as \textit{normalisation factor} and defined as $3V/A$ in \cite{rodriguez-carranza_framework_2008} while in \cite{knutsen_spatial_2013} it was called \textit{characteristic length} and defined as $\sqrt{A/4\pi}$. These two formula are equivalent in the case of a perfect sphere and correspond to its radius. 
In the case of real brains, however, normalising with respect to either the volume or the area of the surface will not be equivalent since surface area and volume are related by an allometric rule, as pointed in the introduction.
In the absence of better understanding of the causal mechanisms relating brain maturation and folding, normalising using either surface are or volume amounts to making an assumption in favour of one theoretical model of brain folding against the others \cite{ronan_genes_2015}.
Our method also requires the estimation of a normalisation factor, we are thus confronted to the same question.
In this work, we used the cubic root of the volume. 
This choice is based on the assumption that brain folding is a consequence of a mechanical instability due to brain growth \cite{tallinen_growth_2016}, which implies that the increase of the surface area is a consequence of the increase in tissues volume. 

To our knowledge, our study is the first to extend the normalisation of brain measures beyond those defined by simple formula, and in particular to partial differential equations. 
The nondimensionalisation approach we used is relatively standard in the field of differential equations coming from the physics \cite{sanchez_perez_searching_2017}.
It implies considering characteristic quantities of the problem, e.g. with respect to space or time. 
In our case we only needed to define a characteristic length. 
An interesting extension of our approach could be to incorporate a characteristic time involved in a spatio-temporal evolution, such as surface smoothing as proposed in \cite{lefevre_surface_2013}.

\subsection{Tuning of the parameter for $DPF^*$}

Like the original $DPF$ method in \cite{boucher_depth_2009}, the $DPF^*$ depends on a parameter $\alpha$ which has the physical unit of the inverse of an area ($m^{-2}$). We showed in section \ref{sec:alpha_setting} the relationship between this parameter and the size of a spatial kernel: a larger parameter value induces smaller spatial neighbourhood influencing the measure and vice versa. In our approach, the scale-control property enables to obtain a similar regularisation level despite variations in brain size. To our opinion it is a crucial point which deserves more attention in neuroimaging where smoothing procedures are rarely used with an adaptive kernels for various brain sizes, except for instance \cite{kim_development_2016}.



\subsection{$DPF*$ does not compensate for variations beyond global scaling}

Our approach is the first to explicitly allow for compensating for the influence of global brain size on local depth estimates.
However, the increasing complexity of cortical folding during development \cite{lefevre_are_2016} or allometric variations with respect to brain size between adults \cite{toro_brain_2008, reardon_normative_2018} suggest that more subtle empirical laws exist between sulcal depth and the global size of the brain. 
Specifically, our observation of a strong influence of brain size (higher Wasserstein distance) on the $DPF*$ for the 200 smaller, younger subjects despite our formal normalization is expected. We interpret the sharp increase in variance for DPF* on this age range as the impact of the combination of increase in global size and magnitude of folding, that cannot be compensated by a global normalization factor.
Regarding SULC, our experience 1 confirms that an increase in global size is sufficient to induce a spread in the distribution of the estimated depth, which is expected from a method that is not normalized. The observation of the increase in variance with age for very young babies is thus expected.

Normalizing for global scaling should thus not fully compensate for variations in sulcal depth.
In this work, we propose an original formalization of the problem that is a contribution to addressing this question, but not in terms of biological processes. Additional studies are needed to better understand the interactions between biological processes and their dynamics, in order to approach the causal relationship between mechanisms acting at the cellular level, and the observed variations at the millimeter to centimeter scale (folds) and finally at the global brain scale.
The ultimate sulcal depth estimation technique would eventually consider local normalization factors. 
Designing such a method allows us to also compensate for local variations in cortical geometry, for instance using the approach introduced in \cite{joshi_parameterization-based_2009}.
However, note that modelling variations in local cortical folding would amount to propose a model of brain folding like for instance \cite{tallinen_growth_2016}.
The inclusion of local normalization factors thus corresponds to a major change in the objectives and perspectives of the approach.

\section{Conclusion}
In this work, we proposed a formalization of the influence of global scaling on functions defined on surfaces.
We applied this formalism to introduce the $DPF^*$, a sulcal depth estimation technique involving an explicit normalization with respect to the global size of the brain.
We demonstrated the relevance and efficiency of this new method in both simulated and real data.

\section*{Ethics statement}
This research study was conducted retrospectively using human subject data made available in open access by the dHCP (http://www.developingconnectome.org), the HCP (https://www.humanconnectome.org/) and the KKI (Kennedy Krieger Institute, \cite{landman_multi-parametric_2011}). Each study was approved by the respective institutional review board.
All procedures were performed in compliance with relevant laws and institutional guidelines and have been approved by the appropriate institutional committees. Informed consent was obtained for experimentation with human subjects.

\section*{Acknowledgments}
The HCP data were provided by the Human Connectome Project (1U54MH091657-01) supported by the 16 NIH Institutes and Centers that support the NIH Blueprint for Neuroscience Research and NIH ROI MH-60974.
The dHCP data were provided by the developing Human Connectome Project, KCL-Imperial-Oxford Consortium funded by the European Research Council under the European Union Seventh Framework Programme (FP/2007-2013) / ERC Grant Agreement no. [319456]. We are grateful to the families who generously supported this trial.

We thank the Agence National pour la Recherche for supporting this study under the Sulcal Grid project (ANR-19-CE45-0014) and the Eranet project (ANR-21-NEU2-0005). 

\section*{Declaration of generative AI and AI-assisted technologies in the writing process}
During the preparation of this work the authors used MISTRAL AI in the writing process to improve the readability and language of the manuscript. After using this tool/service, the authors reviewed and edited the content as needed and take full responsibility for the content of the published article.

%
\bibliographystyle{apalike}
\bibliography{main.bib}

\section*{Supplementary Material}

\subsection{Protocol for the manual annotation of specific geometrical features}
\label{app:protocol}

The manual annotation of \textit{anatomical} objects on individual cortical surfaces is an ill-posed problem, since the biological relevance of the traced objects cannot be assessed in vivo.
In the present work, we circumvent this limitation by focusing on \textit{geometrical} objects that can be identified with \textit{minimal ambiguity} by adopting a conservative tracing strategy.
In this annex, we describe the full protocol we designed to define and annotate a set of lines allowing us to evaluate and compare different sulcal depth estimation methods on brains with large variations in global size. 

\subsubsection{Data selection}
We select for the annotation process a set of 17 different subjects with high quality cortical surface mesh; and representative of the range of age and volume of the human brain development. 
The 16 subjects consisted of 13 newborns from the dHCP database  with an age range of [29 GW ,44 GW] and a range of volume of [50000 $mm^3$, 225000 $mm^3$]; and 3 young adults from the KKI database with a volume of respectively $324891 mm^3$, $468156 mm^3$, $608843 mm^3$.

\subsection{Prototypal sulcal bassin}

We first define the \textbf{prototypal shape of a sulcal basin} consisting of three identifiable and separable regions: 1) the fundus defined as the lower region; 2) the crest defined as the ridge between the depletion and the enclosing upper region; and 3) the sulcal wall defined as the relatively planar region located between the upper and lower parts of the sulcal basin. 
From these three regions, we can define the three types of line of interest corresponding to key geometrical features of a sulcal basin : 
\begin{itemize}
    \item The \textbf{gyral crest} correspond to the lines surrounding the sulcal basin, following the local mean curvature maxima between the wall of the basin and the surrounding gyral regions.
    We choose to trace the lines corresponding to the lip surrounding the basin instead of the lines that would be obtained by a morphological skeletonisation of gyri, because that alternative definition is not appropriate for smoother brains corresponding to young babies showing large flat or slightly convex regions for which the skeleton would not be relevant.
    \item The \textbf{fundus} corresponds to the bottom of the sulcal basin, defined as the location of the maximum local mean curvature in the transverse plane. Along the longitudinal plane, we trace the fundus between the two inflection points that separate the walls from the bottom of the basin. 
    \item The \textbf{directional lines} correspond to a set of shortest geodesic path between the crest and the fundus from a given sulcal basin.
\end{itemize}  

We then define below the range of \textbf{acceptable deviations}, i.e. we limit the tracing to cortical folds for which the identification of each of the three regions and corresponding lines can be traced without ambiguity.
The principle is to be conservative in the choice of the basins and lines we trace on a given cortical mesh by selecting only the sulcal basins for which we can identify the three regions and corresponding lines of interest.
In other words, we prefer to trace a small number of lines with low ambiguity than increasing the number of lines by including ambiguous geometric configurations.

For each subject, we start with the simplest, less ambiguous basin and then iteratively work our way to more complex ones. 
We stop tracing when there is no more unambiguous basin to trace. 
A crucial aspect, explained in more detail below, is the set of criteria that lead to the decision that a basin is too ambiguous to be drawn. 

\subsection{Range of acceptable deviations from the prototypal configuration}
\begin{figure}[h!]
\includegraphics[width=\linewidth]{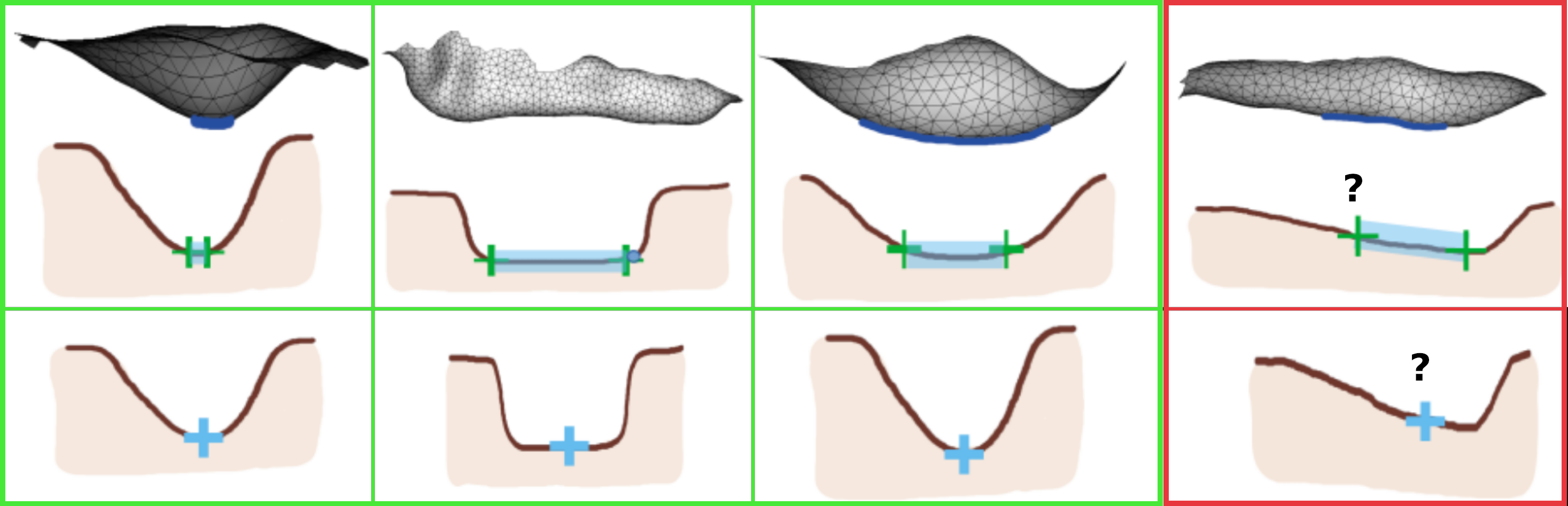}
\caption{Illustration of acceptable and unacceptable deviations with respect to the prototypal sulcal basin.
Top: longitudinal view. Bottom: transverse view. The three examples of the left illustrate acceptable variations, while the example on the right shows ambiguous configurations. The basin on the right should not be considered for manual annotation.}
\label{fig:supp_acceptable_deviations} 
\end{figure}

We illustrated on Fig.\ref{fig:supp_acceptable_deviations} and Fig.\ref{fig:supp_conservative_tracing} the deviations from the prototypal sulcal basin and lines we describe in this section.
We define the range of acceptable deviations following several dimensions:
\begin{itemize}
    \item The size. The acceptable basins correspond to a marked concavity with a clear depletion. Note that since the overall procedure imposes starting from less ambiguous basins before moving to more ambiguous, smaller ones, this criterion will be decisive mostly for younger, smoother brains.
    \item Deviations along the longitudinal plane (tortuosity, intersections, flattening):
    In the case of basins with intersections such as a Y shape, we consider the different branches as different basins for which we apply the criteria independently. 
    In case of local elevation of the fundus resulting from a buried gyrus, the basin is considered split and the two subparts are considered independent basins.
    A crucial aspect is the identification of the end points of the fundus.
    \item Deviations along the transverse plane (opercularization): variations in shape that do not induce ambiguity in the identification of the fundus of the basin are accepted, but as soon as the identification of the fundus becomes unclear, do not trace. In particular, we do not trace for basins with widening inducing large flat fundus, or opercularization inducing a marked asymmetry. Because of this criterion, we do not trace lines surrounding the insula. 
    \item Deviations of crest lines (sulcus opening and buried gyrus): We stop the tracing of the crest line when there is a drop along the crest in the direction of the bottom, even if it corresponds to a slight depression. 
\end{itemize}  

\subsection{Manual annotation of fundi and gyral crests}

Regarding the concrete tracing of lines, note that using our protocol, the tracing of the different lines corresponding to a particular basin is simplified by the decision of not tracing ambiguous geometrical configurations. To further enforce reliable tracing, we adopt the following general procedure:
1) The tracing begins with the fundus: start from the less ambiguous portion of the fundus, and then follow the fundus as long as it stays clearly marked.
No constraint on the topology of the fundus line, it can be split in several pieces e.g. in case of a buried gyrus.
2) Then trace the gyral crests associated to each fundus line. No constraint on the topology of gyral crest, i.e. do not enforce a continuous line surrounding the basin.

\begin{figure}[h!]
\includegraphics[width=\linewidth]{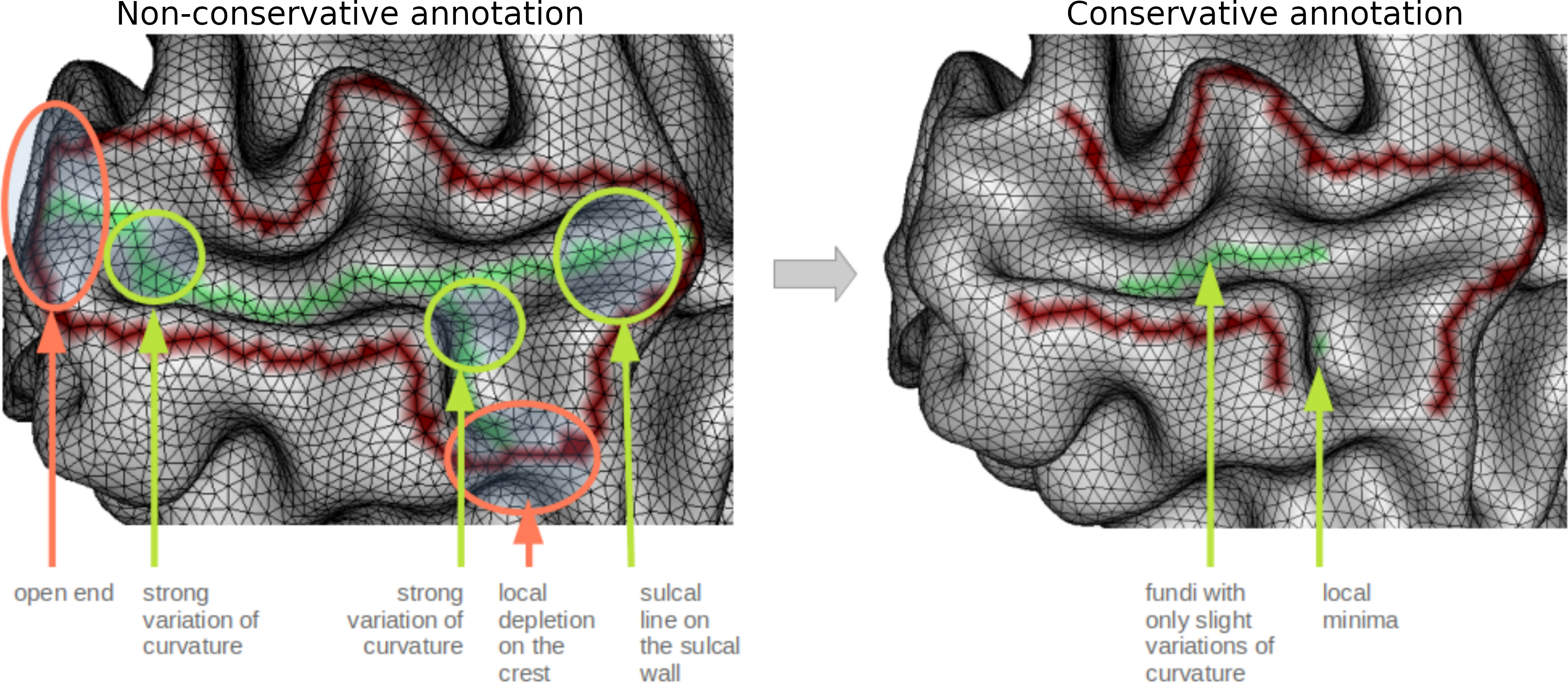}
\caption{Illustration of conservative annotation approach proposed in this protocol.}
\label{fig:supp_conservative_tracing} 
\end{figure}

\begin{figure}[h!]
\includegraphics[width=\linewidth]{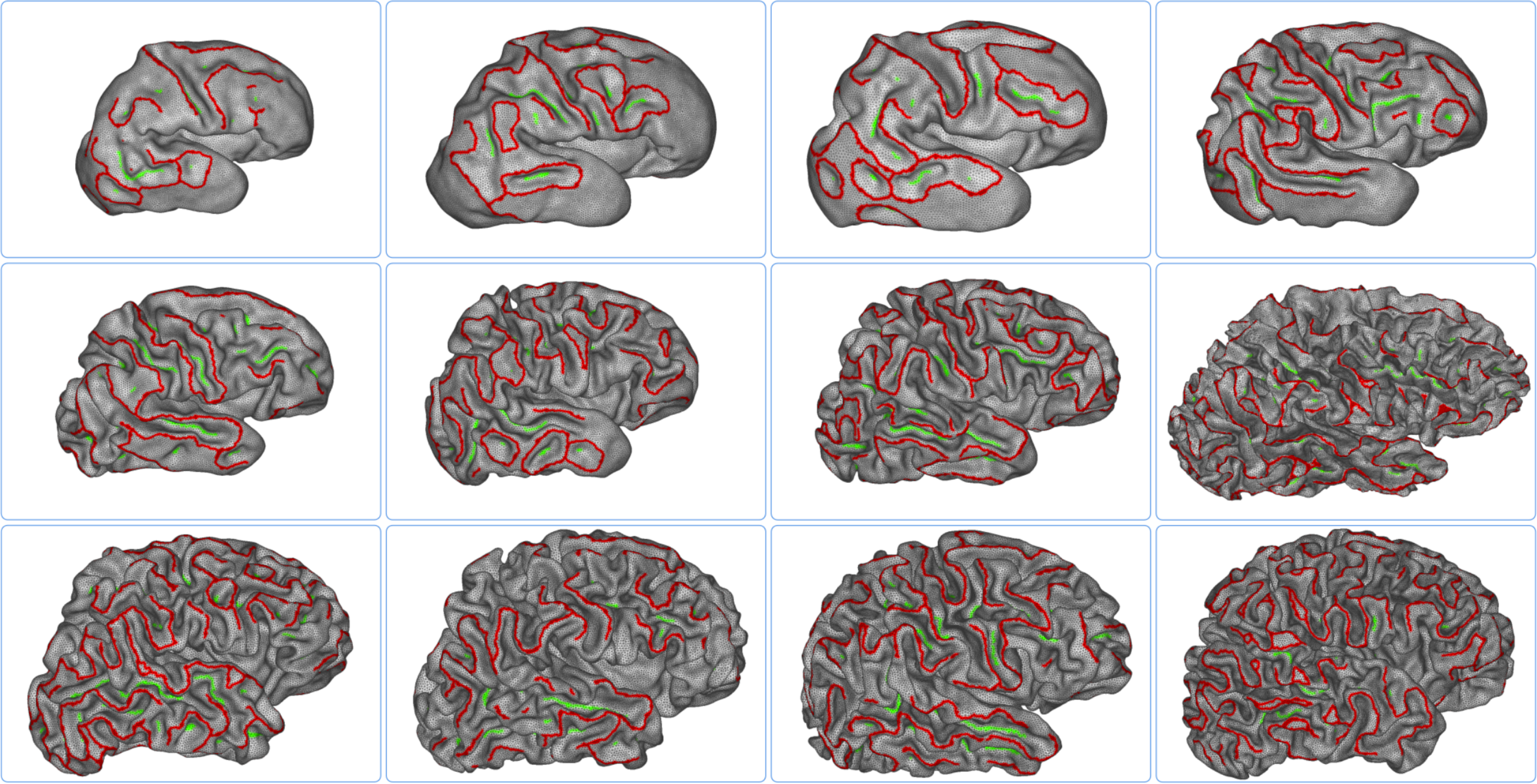}
\caption{Illustration of the annotated fundi (in green) and gyral crests (in red) obtained for a representative sample of cortical surfaces using the proposed protocol.}
\label{fig:supp_tracing} 
\end{figure}

We do not impose any explicit constraint on the relative length of fundus VS crest lines of a given basin.
Depending on the variations in the shape of basins, it might happen that a basin has a very short fundus line and a much longer crest or the reverse.
We provide on Fig.\ref{fig:supp_conservative_tracing} an illustration of the conservative annotation approach we propose.
We also provide on Fig.\ref{fig:supp_tracing} an illustration of the manual annotation obtained for a representative sample of surfaces of various size and magnitude of folding.

\subsection{Automated annotation of directional lines}
The directional lines are the set of shortest paths between the gyral crests and fundus lines of a given sulcal basin. 
Once the crests and the fundi have been annotated, we automatically generate direction lines across the entire surface of each brain. 
Our algorithm consists of the following steps : 1) For each vertex on the fundus, compute the shortest geodesic path to the closest vertex on a gyral crest; 2) For each vertex on the gyral crests, compute the shortest geodesic path to the closest vertex on a fundus line; 3) keep as direction lines the set of shortest path that are common to steps 1) and 2), i.e. the paths corresponding to the shortest geodesic path going both ways.
We then validate these lines by visual inspection.

\subsection{Qualitative comparison on a spatio-temporal atlas}

\begin{figure}[h!]
\includegraphics[width=\linewidth]{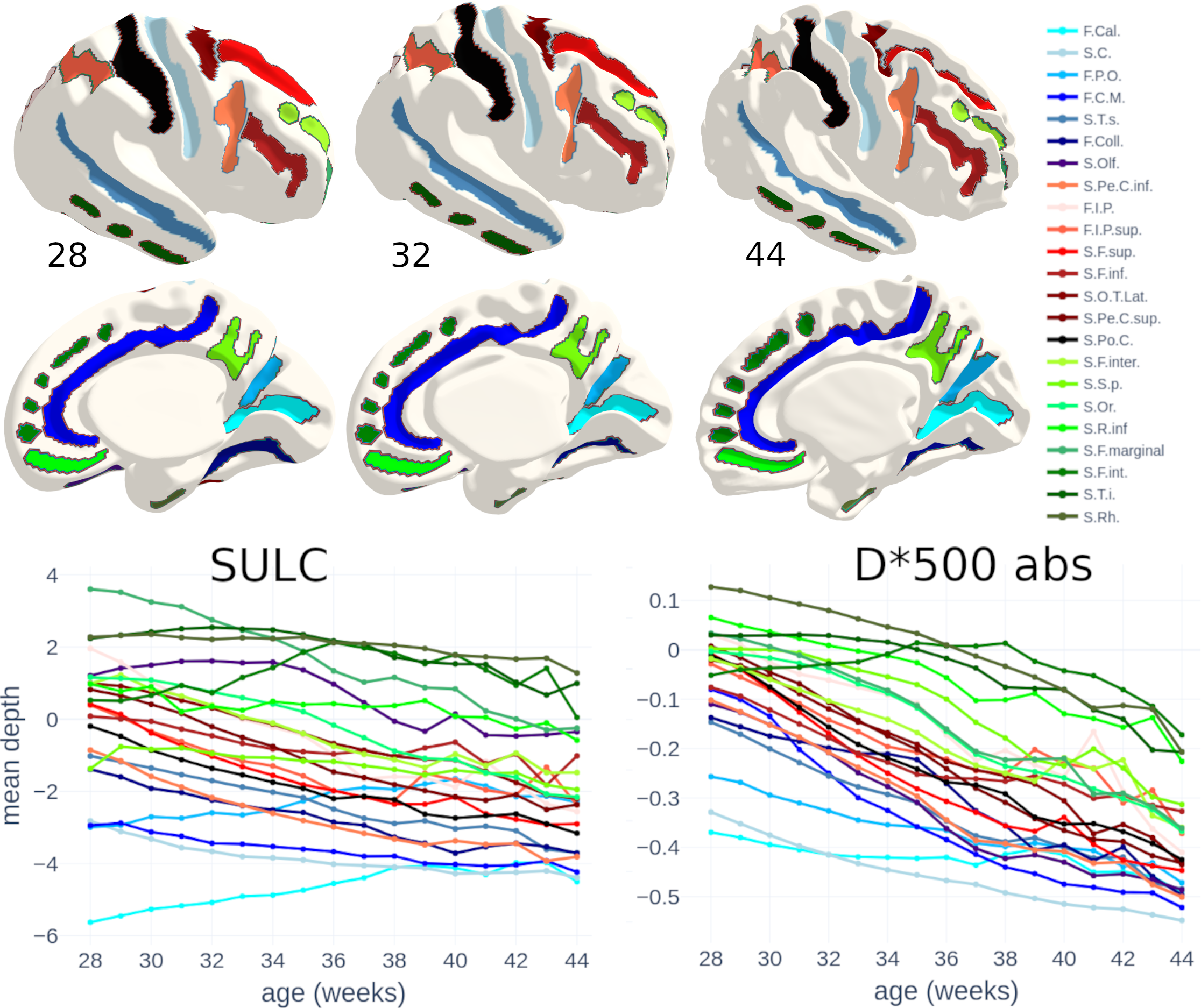}
\caption{Results from Experiment \#4 for the right hemisphere.
Top: Labeled sulcal regions shown on three time steps of the spatio-temporal atlas.
The nomenclature of sulci and associated colors are shown on the right.
The colors are indicative of the expected timing of formation: sulci colored in levels of blue are expected to appear before the sulci in red that appear before the sulci in green.
Bottom: Curves showing the average sulcal depth estimated in the different sulcal regions for SULC (left) and $DPF_{abs}^*$ (Right).}
\label{fig:atlas_sulci_right} 
\end{figure}

We provide on Fig.\ref{fig:atlas_sulci_right} the results from Experiment\#4 for the right hemisphere.
The observations and interpretations are similar to these reported for the left hemisphere in section \ref{sec:res_expe4}

\begin{figure}[h!]
\centering
\includegraphics[width=0.6\linewidth]{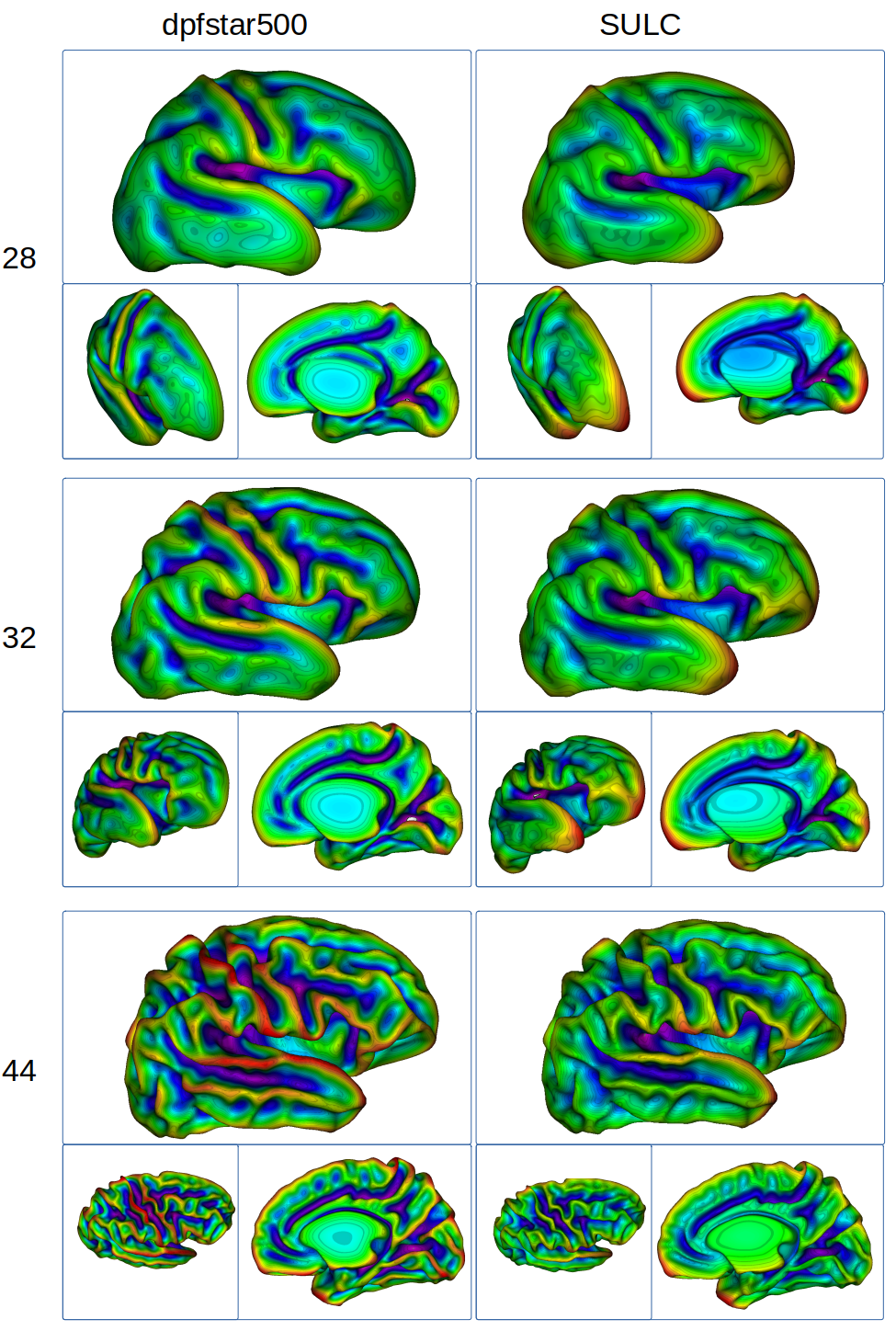}
\caption{Application of $DPF^*$ (left) and SULC (right) to the spatio-temporal atlas from \cite{bozek_construction_2018}, computed from the dHCP dataset corresponding to 28, 32 and 44 weeks post-conception from top to bottom.}
\label{fig:dHCP_template} 
\end{figure}

In order to complement the results and observations from the Experiment\#4, we show in Fig. \ref{fig:dHCP_template} the sulcal depth maps obtained using $DPF^*$ and SULC on the surfacic atlas from the dHCP \cite{bozek_construction_2018} at 28, 32 and 44 weeks post conception.
We observe a better dissociation between sulci and gyri at every age for $DPF^*$ compared to SULC.
In addition, the estimated depth values are more homogeneous in gyral crests distributed across cortical regions with $DPF^*$, while with SULC we observe that high depth values are concentrated around large convex regions such as the anterior temporal lobe, the frontal pole and the occipital region.
The $DPF^*$ gives a finer description of the folds, with a better definition of the small sulci located in the inferior temporal region, as well as in the medial side of the cortex (see for instance the small folds above the cingular sulcus).

\subsection{Influence of the spatial resolution of the mesh}

\begin{figure}[h!]
\centering
\includegraphics[width=\linewidth]{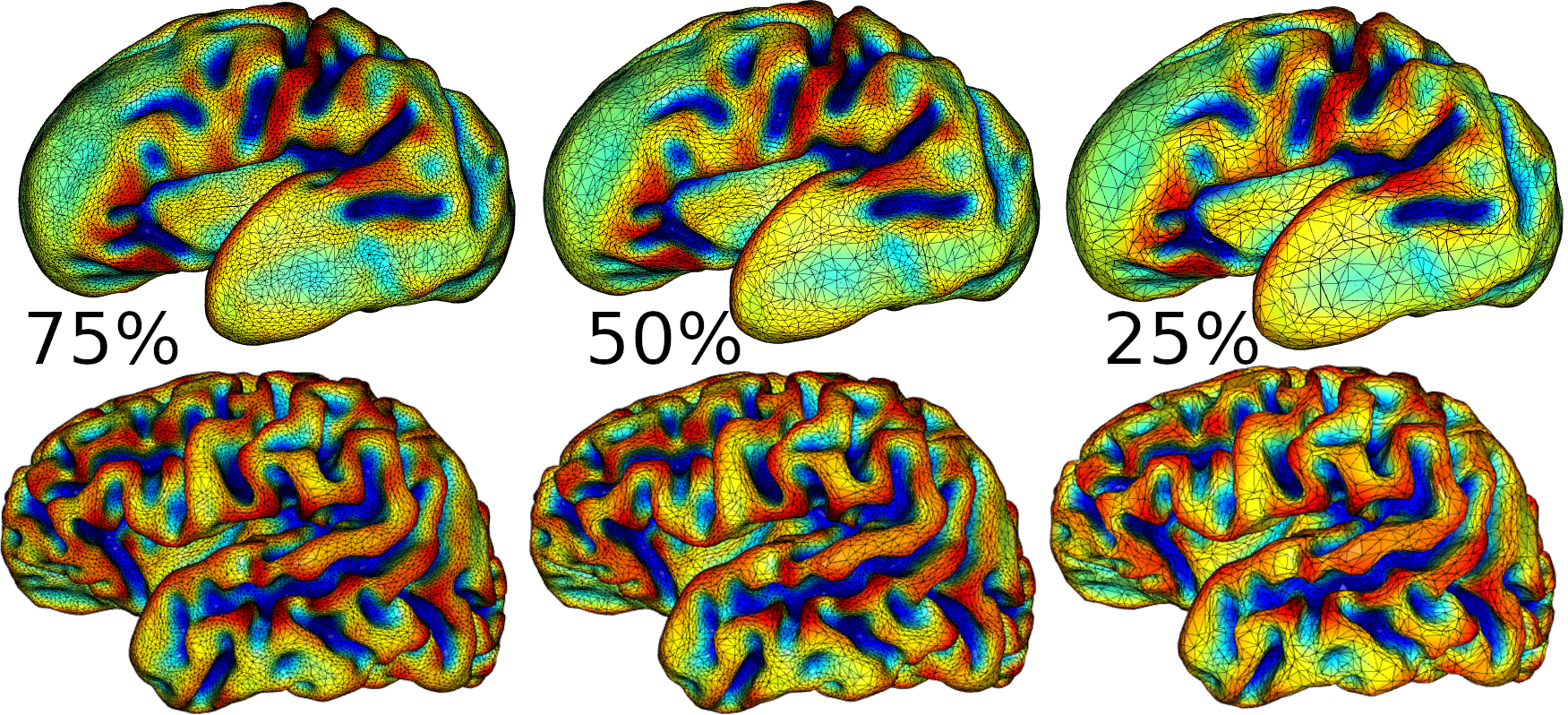}
\caption{Illustration of the $DPF^*$ computed for two individual surfaces after down sampling the number of vertices to 75\%, 50\%, and 25\% of the original mesh.}
\label{fig:exp5_resolution} 
\end{figure}

The impact of the spatial resolution on the sulcal depth value estimated with $DPF^*$ is limited. 
To complement the quantitative assessment from Table \ref{tab:resolution}, we provide on Fig.\ref{fig:exp5_resolution} a visualization of the sulcal depth maps obtained after reducing the number of vertices to 75\%, 50\%, 25\% and 10\% for two subjects with different levels of folding.

\end{document}